%% file: main.tex
\documentclass{article}

\usepackage[final]{neurips_2022}

\usepackage[utf8]{inputenc} % allow utf-8 input
\usepackage[T1]{fontenc}    % use 8-bit T1 fonts
\usepackage{hyperref}       % hyperlinks
\usepackage{url}            % simple URL typesetting
\usepackage{booktabs}       % professional-quality tables
\usepackage{amsfonts}       % blackboard math symbols
\usepackage{nicefrac}       % compact symbols for 1/2, etc.
\usepackage{microtype}      % microtypography
\usepackage{xcolor}         % colors
\usepackage{wrapfig}

\title{Understanding Hyperdimensional Computing for Parallel Single-Pass Learning}

\author{%
  Tao Yu\thanks{Equal Contribution.} \\
  Cornell University \\
  \texttt{tyu@cs.cornell.edu} \\
  \And
  Yichi Zhang\footnotemark[1] \\
  Cornell University \\
  \texttt{yz2499@cornell.edu} \\
  \AND
  Zhiru Zhang\\
  Cornell University \\
  \texttt{zhiruz@cornell.edu} \\
  \And
  Christopher De Sa \\
  Cornell University \\
  \texttt{cdesa@cs.cornell.edu} \\
}

\input{macro}  % self-defined macros
\begin{document}

\maketitle

\input{sections/abstract}
\input{sections/introduction}
\input{sections/related}
\input{sections/preliminary}
\input{sections/limits}
\input{sections/rff-encoding}
\input{sections/groupvsa}

\input{sections/sgd}
\input{sections/experiments}
\input{sections/conclusion}
\input{sections/acknowledgement}
% \section*{References}
\bibliography{reference}
\bibliographystyle{plainnat}

%%%%%%%%%%%%%%%%%%%%%%%%%%%%%%%%%%%%%%%%%%%%%%%%%%%%%%%%%%%%
\input{sections/checklist}

%%%%%%%%%%%%%%%%%%%%%%%%%%%%%%%%%%%%%%%%%%%%%%%%%%%%%%%%%%%%
\appendix
\input{appendix_sections/Proofs}
\input{appendix_sections/GroupLearning}

\end{document}

%% file: macro.tex
\usepackage{algorithm}
\usepackage{algorithmic}
\usepackage{enumitem}
\usepackage{multirow}
\usepackage{amsthm}

\newenvironment{tightlist}[4][\textbullet]
{
  \begin{list}{#1}
  {
    \setlength{\leftmargin}{#2}
    \setlength{\rightmargin}{#3}
    \setlength{\topsep}{0pt}
    \setlength{\parsep}{0pt}
    \setlength{\listparindent}{0pt}
    \setlength{\itemsep}{#4}
  }
}{
  \end{list}
}

\usepackage{amsmath,amsthm,amssymb}
\usepackage{mathtools}

\newtheorem{statement}{Statement}[section]
\newtheorem{theorem}{Theorem}[section]

\newtheorem{lemma}[theorem]{Lemma}

\newtheorem{definition}{Definition}

\newcommand{\R}{\ensuremath{\mathbb{R}}}

\renewcommand{\epsilon}{\varepsilon}

\DeclareMathOperator*{\expect}{\mathbb{E}}
\DeclareMathOperator*{\sgn}{sgn}
\DeclareMathOperator*{\prob}{\mathbb{P}}

\renewcommand{\vec}[1]{\ensuremath{\boldsymbol{#1}}}
\newcommand{\mat}[1]{\ensuremath{\mathbf{#1}}}

\newcommand{\simi}{\ensuremath{\mathcal{S}}}

\newcommand{\bb}[1]{\mathbb{#1}}

\newtheorem{innercustomthm}{Theorem}
\newenvironment{customthm}[1]
  {\renewcommand\theinnercustomthm{#1}\innercustomthm}
  {\endinnercustomthm}

\newtheorem{innercustomlem}{Lemma}
\newenvironment{customlem}[1]
  {\renewcommand\theinnercustomlem{#1}\innercustomlem}
  {\endinnercustomlem}

\newtheorem{innercustomstate}{Statement}
\newenvironment{customstate}[1]
  {\renewcommand\theinnercustomstate{#1}\innercustomstate}
  {\endinnercustomstate}

%% file: sections/abstract.tex
\begin{abstract}
Hyperdimensional computing (HDC) is an emerging learning paradigm that computes with high dimensional binary vectors. There is an active line of research on HDC in the community of emerging hardware because of its energy efficiency and ultra-low latency---but HDC suffers from low model accuracy, with little theoretical understanding of what limits its performance. We propose a new theoretical analysis of the limits of HDC via a consideration of what similarity matrices can be ``expressed'' by binary vectors, and we show how the limits of HDC can be approached using random Fourier features (RFF). We extend our analysis to the more general class of vector symbolic architectures (VSA), which compute with high-dimensional vectors (hypervectors) that are not necessarily binary. We propose a new class of VSAs, finite group VSAs, which surpass the limits of HDC. Using representation theory, we characterize which similarity matrices can be ``expressed'' by finite group VSA hypervectors, and we show how these VSAs can be constructed. Experimental results show that our RFF method and group VSA can both outperform the state-of-the-art HDC model by up to 7.6\% while maintaining hardware efficiency. This work aims to inspire a future interest on HDC in the ML community and connect to the hardware community.
\end{abstract}

% Important to study this method theoretically, because people are actually building hardware then implement these methods, and the appropriate people to review that type of work are people in the ml community, we hope this paper will start some level of interest in the ML community and connect to the hardware community, to make sure what they are doing is sensible theoretically

%% file: sections/introduction.tex
\section{Introduction}
% no \IEEEPARstart

% Problem 1: limitations of 1-bit VSA

% 1. HDC/VSA is an emerging brain-inspired learning paradigm. It is highly compute efficient. \zz{cite papers from hardware (HW) confs. HDC is projected to be very efficient in emerging HW, not necessarily on commodity HW.}

Hyperdimensional computing (HDC) is an emerging learning paradigm. 
Unlike conventional cognitive modeling that computes with real numbers, it computes with high dimensional \emph{binary} vectors, referred to as binary \emph{hypervectors}, the dimension of which is usually at least in the thousands.
% , referred to as hypervectors. 
HDC is brain-inspired as high dimensional representations have two fundamental properties similar to human brains: they are 
\textbf{(1)} distributed and highly parallel; and
\textbf{(2)} robust to noise and tolerant to component failure~\citep{Kanerva2009hdc}.
On the other hand, the massive parallelism and simple arithmetic project HDC into the scope of energy-efficient and ultra-low-latency computing, especially with the rise of emerging hardware~\citep{imani2021revisit, imani2020clustering, gupta2020thrifty, salamat2019f5hd}.
As a result, HDC has recently attracted considerable attention from edge applications, e.g., robotics, DNA pattern matching, and health diagnosis, as well as data center applications such as recommendation systems~\citep{mitrokhin2019hyperperception, neubert2019hdcRobotics, neubert2021hyperdimensional, kim2020dna, burrello2019seizure, guo2021hyperrec}.
% \fixme{(cite applications such as EEG, robotics, rec sys?)}.
% \zz{we need to mention ``single-pass learning'' somewhere in intro or prelim to explain the title}

% 2. The problem is: Compared to other alternative learning methodology --- such as neural networks, HDC presents a lower accuracy.

The practical deployment of HDC is undermined by its low model accuracy compared to other alternatives, e.g., neural networks (NN). 
The state-of-the-art HDC model on MNIST has an accuracy of $89$\%~\citep{chuang2020dynamicHDC}. 
A two-layer NN, however, can easily achieve $95$\%~\citep{lecun1998recognition}.

% 3. The current approach of improving HDC model accuracy focuses on increasing the dimensionality of the hyper vectors. However, we find that the ability of achieving a certain similarity matrix is essential to improving the model capacity.

There are two main approaches in the literature to improving HDC.
One approach is to increase the hypervector dimension, staying within the classic HDC paradigm and just making the binary vectors longer \citep{neubert2019hdcRobotics, schlegel2021vsa}.
An alternative is to increase the complexity of each element in a hypervector, e.g., to floating-point or complex numbers (unit circle in the complex plane) \citep{plate1995holographic, gallant2013represent, Gayler1998MultiplicativeBR, plate1994complexHDC}: this moves the system into the more general realm of \emph{vector symbolic architecture} (VSA) \citep{schlegel2021vsa}, which uses high-dimensional vectors with elements that are not necessarily binary (unlike binary HDC). 
However, these remedies are not based on any theoretical analysis of the limits of HDC, and as a result there is a lack of more-than-empirical understanding of when and how they should be deployed.

% \textcolor{ForestGreen}{This paragraph should describe our similarity matrix contribution briefly---point out that this notion of express-ability distinguishes between larger-vector and more-complicated-vector approaches.}
% \textbf{We do not understand more than empirically the limitations of binary hypervectors.} With a fixed set of arithmetic, it is unknown where the expressivity of binary hypervectors is from, and whether there exists a scenario where increasing the dimensionality even to infinity provides no quality gain. \tao{maybe we can provide some empirical examples where increasing the dimensionality will not help improve the model accuracy, then we can change this statement a bit.}
% 1. Introduce the notion of similarity matrix as the express-ability
% 2. Concequence is that express-ability is not a function of vector length in VSA
% 3. So choosing a specific length of precision of VSA over another is independent of 
In this work, we introduce a new notion of expressivity for any VSA using \emph{similarity matrices}. 
Given a set of hypervectors $v_1, \ldots, v_n$ in a VSA, each entry $\mat{M}_{ij}$ in the similarity matrix $\mat{M}$ is defined as the similarity between a pair of hypervectors $v_i$ and $v_j$; the similarity is typically measured by an inner product function.
Informally, we propose that a VSA is \emph{more expressive} when it can express (i.e. represent with some set of vectors) a wider class of similarity matrices. 
% We find that instead of a function of vector dimension or complexity of elements in the hypervector, the expressivity of a general VSA is highly correlated to what similarity matrix it can express.
Importantly, which $\mat{M}$ a VSA can express is independent of the vector dimension $D$: this new notion distinguishes between longer-vector and more-complex-vector approaches.
%It provides a mechanism about how the expressivity will change when the dimension of vector or component complexity varies.
% either varying the vector length or component complexity.

% \textcolor{ForestGreen}{This paragraph should describe our group VSA approach and connection to representation theory.}
% \textbf{There is also a missing methodology on interpolating between HDC and a general VSA that strives for a better accuracy but still preserves the computation efficiency.} Though having better model qualities, some VSAs employing floating-point or complex number hypervectors \fixme{cite} require sophisticated compute units on hardware. They are less suitable for edge applications with a limited budget on power and circuit area \fixme{cite the applications in the first paragraph again}.

We show that HDC, with binary hypervectors even in \emph{any} dimension, cannot express as many similarity matrices that a VSA with more complex hypervectors can.
Even worse, the current method of initializing the hypervectors in an HDC system further reduces the expressible set, which impedes the success of HDC.
This notion of expressivity is closely related to learning ability. We exhibit a simple task where current HDC (of any $D$) is incapable of learning a Bayes optimal classifier, while any other VSA system that can express a particular similarity matrix $\mat{M}$ (which HDC cannot express) can learn it through the same procedure.

Based on our analysis, we investigate how we can improve HDC through the lens of similarity matrices.
We first propose to improve the initialization of binary hypervectors by employing random Fourier features (RFFs) \citep{rahimi2007rff}. 
This method is different from existing approaches that increase the dimension or complexity of hypervectors.
We show that this better initialization via expressing a similarity matrix can already surpass state-of-the-art HDC accuracy on MNIST by $6.4$\%.
We then propose and formally define group VSA, an extended version of HDC where elements in hypervectors are more complex than binary but less than floating-point.
Group VSA can further improve the RFF-initialized HDC by $1.2$\% on MNIST.

% In this work, we show the theoretical limitations of HDC based on representation theory. Our analysis then inspires a generalization of HDC to group VSA that bypasses the limitations. Specifically, we demonstrate an example where the current HDC system with binary hypervectors are incapable of learning a simple task. This promotes the importance of the \emph{similarity matrix} that will be described in Section~\ref{sec:hd-arithmetic} and Section~\ref{sec:limits}. We then use random Fourier features \fixme{cite}, a typical method for approximating a similarity matrix, to improve the current HDC system as a proof of concept \tao{some numbers on MNIST as comparison to SOTA}. In Section \ref{}, we provide group VSA as an interpolation between binary HDC and complex VSA with formal definitions, which has better expressivity while still preserving the computational efficiency. We also propose a new way to better learn via VSA rather than using commonly adopted bundling method in Section \ref{}. Experiments results of our method is provided in Section \ref{}.

% 5. List of Contributions.

Our contributions are as follows:
\begin{tightlist}{1em}{0.3em}{0.3em}
    \item We provide a theoretical analysis of the limitations of an HDC system with binary hypervectors.
    \item We approach the expressivity limit of HDC systems with random Fourier features and empirically evaluate the improvements on standard benchmarks. 
    \item We propose group VSA, which generalizes HDC with more complex elements, expanding the set of expressible similarity matrices while maintaining efficiency.
    \item We evaluate the performance of group VSA on both conventional HDC tasks and image tasks, and study its efficiency implications by analyzing the circuit depths.
\end{tightlist}

% \tao{story: finite HDC and complex version of HDC exists, unknown: 1. limitations of acting binary, only empirically. 2. missing a way to interpolate to produce an intermediate method between the extreme binary and more general VSA, trading off system and statistical considerations, there problems are related, the fact we don't understand the theoretical limitation of binary case is because we don't really understand how to construct something that bypass these limitations, while still keeping computational efficiency and accuracy. In this paper, with a novel analysis based on representation theory, we are able to solve both of the problems, we show the limitations of binary HDC in section xxx, and we show how to bypass these limitations with a natural generation of HDC to groups. In our setup, HDC corresponds to group of two elements, and the complex number VSA corresponds to the unit cycle group, and by choosing a finite group between them, we can interpolate between them ...}

%% file: sections/related.tex
\section{Related Work}

The term HDC was first introduced by \citet{Kanerva2009hdc}. It is also referred to as VSA in some literature \citep{schlegel2021vsa}, a line of work that does symbolic computing.
Binary HDC can be traced back to Binary Spatter Code (BSC) \citep{kanerva1994spatter,kanerva1997fully}. 
% It can be traced back to a branch of VSA 

\textbf{Model capacity improvement.}
There are two main VSA formats other than binary HDC: using floating-point real vectors \citep{plate1995holographic, gallant2013represent, Gayler1998MultiplicativeBR, gosmann2019vtb} or complex vectors \citep{plate1994complexHDC}.
Typically their model capacities are higher since their individual vector components are more complex.
Another way of increasing the model capacity is increasing the vector length \citep{neubert2019hdcRobotics, schlegel2021vsa, chuang2020dynamicHDC,frady2018theory}.
However, it remains unknown when and how we should apply these methods, and whether they are sufficient to solve a task.
Our approach is different as our proposed methods are based on the theoretical analysis of, and are designed to bypass, the limits of binary HDC.

\textbf{Hardware implication.}
HDC inspires a novel hardware architecture that requires \emph{associative memory} \citep{Hopfield1982memory} where long vectors can be stored and addressed efficiently.
It is therefore popularized recently in the emerging in-memory computing community \citep{imani2020clustering, gupta2020thrifty}.
In the meantime, the simplicity of HDC arithmetic and the massive parallelism make HDC suitable for tasks that require high energy efficiency and low latency.
It has been demonstrated successful on commercial hardware as well \citep{imani2021revisit, salamat2019f5hd, basaklar2021edge}.
In this work, we provide an analysis on the circuit depths of HDC and our proposed group VSA.

\textbf{Theory.}
Understanding HDC from a theoretical perspective is currently limited. \citet{thomas2020theoretical,frady2018theory} presented some theoretical foundations of HDC, introducing the benefit of high-dimensional vectors, hypervector encoding, and the connection between HDC and kernel approximation. 
Our work instead presents the limits of HDC and how we can bypass it.
\citet{frady2021computing} propose to generalize VSA/HDC to function space. Our work is different since our proposed group VSA, a generalization of HDC, is still discrete and preserves the hardware efficiency.

%% file: sections/preliminary.tex
\section{Background on HDC}
\label{sec:background}
% \fixme{Yichi: Shorten it.}
% \tao{1. element of VSA; 2. binding; 3. bundling; 4. similarity measure. basic definition of VSA, most paper use without formal definitions, we are gonna provide a formal definition in section 5.}
% \tao{define what similarity function is and give some properties of it, later formalize more explicitly in section 5., define what similarity matrix is in terms of similarity function, the set of representable similarity matrix for VSA is defined as the closure of the set of matrices that can be represented by a set of vectors of some length in the VSA, ration number for binary HDC.}
In this section we introduce basics of HDC: hypervectors, arithmetic, and the learning paradigm. We then present a classical approach of HDC on the popular MNIST database. A more comprehensive introduction is in \citet{ge2020hdcreview}.

\textbf{HD Representations.}
In HDC, we compute with binary hypervectors in a high-dimensional space referred to as \emph{hyperspace}. 
% Given a random hypervector in a 10,000 dimensional space $v \in \{ -1, 1 \}^{10,000}$, it is well known from the ``curse'' of dimensionality that almost all the vectors in this hyperspace are located in a 600-bit wide bulge that is centered 5000-bit away from $v$. 
Given a random hypervector $v$ in a $10,000$ dimensional space $\{ -1, 1 \}^{10,000}$, it is well known from the ``curse'' of dimensionality that most vectors in this hyperspace are nearly orthogonal to $v$ \citep{Kanerva2009hdc}.
% In other words, the entire hyperspace is nearly 5000-bit away from $v$ \citep{Kanerva2009hdc}.
We call such hypervectors \emph{unrelated}. The ``curse'' provides two intriguing properties for cognitive tasks: (1) independent random hypervectors will be unrelated and so can naturally represent objects that are semantically separate, e.g. letters of the alphabet; (2) two hypervectors $u$ and $v$ that have a high-enough inner-product similarity can be classified as being related (i.e. somehow dependent) with high probability. Classical HDC therefore represents data using binary hypervectors randomly drawn from a hyperspace.
%  \textbf{\textcolor{red}{re-work this paragraph to be less vague, quasi-orthogonal property and cite}}
HDC computes with hypervectors using a fixed set of primitive operations: similarity, binding, bundling, and permutation.

%\paragraph{HD Arithmetic.}
%\label{sec:hd-arithmetic}

%In this section, we present a set of primitive operations in HDC through which data can be encoded into hypervectors. Given a set of hypervectors in the $D$-dimensional hyperspace $\bb{V} = \{ v_1, v_2, \dots, v_n \}, v_{i} \in \{ -1, 1 \}^{D}$, the operations are defined as follows. 

\textbf{Similarity.} A similarity function $\simi(u, v)$ measures how close two hypervectors $u, v \in \{-1,1\}^D$ are. It is typically defined as an inner product function \citep{frady2021computing}
$\simi(u,v) = \frac{1}{D} \sum_{i=1}^D u_i v_i$; this is an affine function of the hamming distance for binary hypervectors~\citep{Kanerva2009hdc}.
% A similarity matrix measures the similarity of each pair of vectors in the given set $\bb{V}$. A formal definition is given in Section~\ref{sec:limits}, where we show that the set of expressible similarity matrices is crucial for the learnability of HDC. 

\textbf{Binding $\bigotimes$.} The binding operation combines two hypervectors $u,v$ into a new hypervector in the same space that represents them as a pair. For binary $\{ -1, +1 \}^{D}$, binding 
% $\otimes$ 
is equivalent to  coordinate-wise multiplication, i.e. $(u \otimes v) \in \{ -1, +1 \}^{D}$ and $(u \otimes v)_i = u_i v_i$ for all $i \in \{1, \ldots, D\}$.
% Binding $v_{1}$ and $v_{2}$ maps $v_{1}$ to another vector in the $D$-dimensional hyperspace. The mapping distance depends on the number of $-1$s in $v_{2}$ as each $-1$ flips a single bit.
Binding preserves similarity, in the sense that $\simi(u \otimes w,v \otimes w) = \simi(u,v)$ for any hypervectors $u$, $v$, and $w$; also, if $u$ is highly similar to $v$ and $x$ is highly similar to $y$, then $u \otimes x$ will be highly similar to $v \otimes y$ (although usually less than either constituent pair). Binding is implemented on hardware as an XOR.
% We can use binding to represent a pair of values.

\textbf{Bundling $\bigoplus$.} Bundling represents an unordered collection of hypervectors. The bundling operation takes in a set of hypervectors and yields a hypervector that is  maximally similar to all of them: it acts as an aggregation of a set of hypervectors. %, resulting in a hypervector that has the same dimensionality and precision.
Bundling $v_1, \ldots, v_m \in \{-1,+1\}^D$ yields
$\left( \bigoplus_{k=1}^m v_k \right)_i = \sgn\left(\sum_{k=1}^m \left( v_k \right)_i \right)
\text{ for all } i \in \{1,\ldots,D\}$.
% We binarize the sum-vector to $\{ -1, 1 \}^{D}$ to preserve the precision, which is essentially taking a majority vote at each coordinate. 
This takes a majority vote at each coordinate of the vector; ties are broken at random.
% Bundling represents an unordered collection of hypervectors. It has an important property --- the restulting bundled vector is \emph{similar} to each addend vector $v_{i}$ \citep{Kanerva2009hdc}. 
HDC typically leverages bundling to learn a class representative.
% This property on similarity makes the bundling operation an effective way of combining examples in a training set that belongs to the same class to generate a class vector.

\textbf{Permutation $\prod$.} The permutation operation is a shuffling of the elements in a hypervector. It can be represented as a multiplication of a permutation matrix $\Pi$. A random permutation on a hypervector yields another hypervector that is unrelated to it. Note that permutation is invertible, meaning that $\Pi^{-1} \Pi v_{i} = v_{i}$. It is thus useful for encoding order and position information. In hardware, permutation usually appears as \emph{shifting} since its implementation is efficient.

\subsection{MNIST as a Case Study}
\label{sec:mnist-case-study}
We outline how to use the classic HDC approach on the MNIST digit recognition task for illustration.

% To illustrate HDC, here we outline how the classic HDC approach would work on the MNIST digit recognition task. 
% As a case study, we introduce the classical approach of HDC to the MNIST dataset.
% MNIST has 50,000 training images and 10,000 test images. Each image is grey-scale and has a spatial resolution of 28$\times$28. 
% A hand-written digit is centered in each image.

\textbf{Encoding}. First, 256 basis hypervectors $\{ v_{0}, v_{1}, \dots, v_{255} \}$ are independently drawn at random from a hyperspace $\{ -1, +1 \}^{D}$: each hypervector $v_i$ represents a pixel intensity $i$. 
%The random hypervectors are now mapped to meaningful entities --- $v_{i}$ corresponds to pixel intensity $i$.
Second, we bind all 784 pixels in an MNIST image by their corresponding hypervectors. Since the binding operation is commutative by definition, but pixels in an image have meaningful relative positions, each pixel hypervector is shifted before joining the encoding to preserve that position information. If the input pixel intensities are $p_0, p_1, \ldots, p_{783}$, then its encoded hypervector $t$ is $v_{p_0} \otimes (\Pi v_{p_1}) \otimes \cdots \otimes ({\Pi}^{783} v_{p_{783}})$, 
% \[
% \textstyle
% t = v_{p_0} \otimes \left( \Pi v_{p_1} \right) \otimes \cdots \otimes \left( {\Pi}^{j} v_{p_j} \right) \otimes \cdots \otimes \left( {\Pi}^{783} v_{p_{783}}\right),
% \]
where ${\Pi}^{j}$ denotes shifting a hypervector $j$ times. \nocite{kleyko2016holographic} % Shifting can be in either direction, but needs to be consistent.

\textbf{Learning}. Encoding yields a set of training hypervectors $\bb{T} = \{ t_{1}, t_{2}, \dots, t_{60,000} \}$.
% We can then derive a class representative for digit. This can be done via bundling all the hypervectors that belong to the same class as an effective way of representing an unordered collection of hypervectors. 
To learn, we bundle all the hypervectors that are from the same digit.
Concretely, a class centroid $s_{c}$ is computed by
$s_{c} = \bigoplus_{i|y_{i}=c} t_{i}$.
Each training image is used only once, making this process \emph{single-pass} learning.

\textbf{Inference}. At the inference time, a given test image is encoded through the same procedure. 
The model outputs the class $c$ with the highest similarity $\simi(t_{\text{test}}, s_c)$.
% The resulting hypervector $t_{\text{test}}$ is then compared to each class vector $s_{c}$. The model outputs the class $c$ with the highest similarity $\simi(t_{\text{test}}, s_c)$.

%% file: sections/limits.tex
\section{Similarity Matrices and the Limits of HDC}
\label{sec:limits}
% \fixme{This section is more about justification of defining similarity matrix as the expressiveness of HDC.}

Traditionally the expressivity of HDC setups is identified with the dimension of the hypervectors $D$. This notion is unhelpful for probing the fundamental limitations of HDC, which do not depend on $D$. In this section, we define a new notion of expressivity, which reveals the limits of HDC. % We further illustrate the limits of HDC by giving a task which no binary HDC is able to learn.
\begin{definition}
\label{def:vsa-express}
An HDC (or VSA) system can \emph{express} a similarity matrix $\mat{M} \in \R^{n \times n}$ if for any $\epsilon > 0$, there exists a $D \in \mathbb{N}$ and $D$-dimensional hypervectors $v_1, v_2, \ldots, v_n$ in the HDC/VSA such that $| \mat{M}_{ij} - \mathcal{S}(v_i, v_j)| \le \epsilon$ where $\mathcal{S}$ denotes the similarity function of the HDC/VSA.
\end{definition}

Informally, this means that the HDC/VSA can approximate $\mat{M}$ arbitrarily well. Limitations on $\mat{M}$ we can express correspond to limitations on the similarity relation we can represent on data: if we have some dataset and
% some idea of 
know how similar each pair of examples 
% in the dataset 
should be, whether or not we can represent the similarity
% our idea
accurately with an HDC embedding depends on whether HDC can express the corresponding $\mat{M}$.
% A more formal and general definition is given in Section \ref{sec:groupvsa}. 
%The choice of basis hypervectors can limit the expressivity of an HDC model. 
% Particularly, we can construct a task such that no binary HDC in any dimension is able to learn, but any VSA that allows to represent a particular similarity matrix is able to learn via the standard bundling method.
% Particularly,
% Perhaps 
Surprisingly, there are some matrices that an HDC system can never express.

\begin{lemma}
\label{lem:sim_hdc_fail}
Binary HDC can not express the matrix 
$\mat{M}_{\textup{Lemma \ref{lem:sim_hdc_fail}}} = \frac{3}{2} I_3 - \frac{1}{2} \mathbf{1}_{3 \times 3}.
% = \left(\begin{smallmatrix}
    %     1 & -\frac{1}{2} & -\frac{1}{2} \\
    %     -\frac{1}{2} & 1 & -\frac{1}{2} \\
    %     -\frac{1}{2} & -\frac{1}{2} & 1 \\
    % \end{smallmatrix} \right).
$
% \[
% \mat{M}_{\textup{Lemma \ref{lem:sim_hdc_fail}}} = \left(\begin{smallmatrix}
%         1 & -\frac{1}{2} & -\frac{1}{2} \\
%         -\frac{1}{2} & 1 & -\frac{1}{2} \\
%         -\frac{1}{2} & -\frac{1}{2} & 1 \\
%     \end{smallmatrix} \right).
% \]
\end{lemma}
Our notion of expressivity corresponds to learning ability, we give an example task for which whether a VSA/HDC approach can learn the Bayes-optimal classifier depends on whether it can express $\mat{M}_{\textup{Lemma \ref{lem:sim_hdc_fail}}}$. Consider a supervised learning task with input set $\mathcal{X}=\{0,1,2\}$, output label set $\mathcal{Y} = \mathcal{X}$, and source distribution $\mathcal{P}(x,y)\!:=\! (1/9 +2p)~\textit{if}~x=y~\textit{else}~(1/9-p)$ 
% \[
% \mathcal{P}(x,y) = \begin{cases} 1/9 + 2p & x = y \\ 1/9 - p & x \ne y \end{cases}
% \]
for some small positive number $p$. 
We say that a VSA can \emph{learn} this task if there exists 
% a $D$ and 
a $D$-dimensional encoding of $\mathcal{X}$ in that VSA such that, when the bundling method in Section~\ref{sec:mnist-case-study} is used on a training set of size $N$ drawn from $\mathcal{P}$, the resulting classifier is the Bayes optimal classifier with arbitrarily high probability as $N$ increases.

\begin{statement}
\label{thm:task_hdc_fail}
Binary HDC cannot learn this task. Any VSA (formalized later in Definition~\ref{def:vsa}) that can express $\mat{M}_{\textup{Lemma \ref{lem:sim_hdc_fail}}}$ can learn this task.
\end{statement}
Details on Statement~\ref{thm:task_hdc_fail}
% and Statement~\ref{thm:task_hdc_succeed} 
are in appendix. This learning task shows that only increasing the dimensionality of hypervectors cannot help learn the correct predictions if unable to express a certain matrix. This implies that our notion of expressivity captures HDC limitations in a way that relates to learning.

\textbf{Limitations due to initialization.}
% Existing literature (\tao{cite}) chooses these basis hypervectors with the goal to create maximally different encodings, so as to robustly distinguish them in the presence of noise or other ambiguities. As a result, hypervectors are usually randomly sampled in the hyperspace, which are likely to be orthogonal to each other with a high probability due to the quasi-orthogonality property of high dimensional space (\tao{cite here}). However, we show later in the section this is not the right method under our metric.
So far in this section we have described limitations that are inherent to using binary representations in a VSA. Classical HDC methods are often limited in an additional way: rather than considering arbitrary binary hypervectors, they use hypervectors that are sampled independently at random. In such a system, any  hypervector used for an embedding (used to represent an entity) is constructed either by (1) independently sampling a binary hypervector where each entry has some probability $p$ of being $1$, or (2) permuting and/or binding some pre-existing hypervectors. Examples of this setup can be found in \citet{burrello2019seizure,smith1990random,imani2019bric}.
Surprisingly, we show that this approach further restricts the set of similarity matrices that can be expressed in expectation.

\begin{lemma}
\label{lemma:classichdc}
Let $u_1, u_2, \ldots, u_K$ be binary vectors sampled coordinate-wise independently at random, where each coordinate of $u_i$ has the same probability $p_i$ of being $1$.
Let $v_0$, $v_1$, and $v_2$ be vectors that result from some composition of binding and permutation operations acting on $u_1, \ldots, u_K$, and let $\mat{M} \in \R^{3 \times 3}$ be their similarity matrix, such that $\mat{M}_{ij} = \simi(v_i, v_j)$. Then
\[
\textstyle
\left\| \expect[\mat{M}] - \left( \begin{smallmatrix}
        1 & -\frac{1}{3} & -\frac{1}{3} \\
        -\frac{1}{3} & 1 & -\frac{1}{3} \\
        -\frac{1}{3} & -\frac{1}{3} & 1 \\
    \end{smallmatrix} \right) \right\|_F \ge \frac{\sqrt{2}}{3},
\]
but this target matrix \emph{can be} expressed by binary HDC.
\end{lemma}
% \begin{proof}
% It is straightforward to show that \textcolor{red}{[add proof]} for some $x, y, z \in [-1,1]$
% \[
% \mathbf{E}[\mat{M}] = \begin{pmatrix}
%         1 & xy & xz \\
%         xy & 1 & yz \\
%         xz & yz & 1 \\
%     \end{pmatrix}.
% \]
% But since $(xy) \cdot (xz) \cdot (yz) = x^2 y^2 z^2$ is a square number, it follows that the upper-triangular elements cannot all be negative. At least one of them must be non-negative, from which the result immediately follows. A binary HDC that achieves this matrix is: $(-1,1,1)$, $(1,-1,1)$, $(1,1,-1)$.
% \end{proof}

%% file: sections/rff-encoding.tex
\section{Encoding Hypervectors via RFF}
\label{sec:rff}
% \fixme{Yichi: Can we delete the ``convecntional'' sentences, as they are not necessarily true.}
% Conventional HDC approaches \tao{cite} do not use bundling to construct embedding, rather with binding, shifting or sampling a new independent vector. This setup restricts the expressivity of the binary HDC. 
% \tao{reorganize words}
% One could try to get around of this issue by using bundling, which is being explored by some recent efforts \tao{cite}. However, 
Our analysis using similarity matrices provides a strong motivation for using more principled methods to construct hypervectors.
% rather than using operators such as binding and bundling. 
%which do not offer any theoretical guarantees. 
We argue that, if there is some similarity matrix $\mat{M}$ we want to achieve, we should directly instantiate hypervectors to match it in expectation.

\begin{wrapfigure}{R}{0.53\textwidth}
\vspace{-2.3em}
\begin{minipage}{0.53\textwidth}
    \begin{algorithm}[H]
      \caption{Construct correlated hypervectors}
      \label{alg:gaussianhv}
    \begin{algorithmic}
      \STATE \textbf{input:} similarity matrix $\mat{M}\in \R^{n \times n}$, dimension $d$
      \STATE \textbf{let} $\hat \Sigma = \sin( \frac{\pi}{2} \mat{M} )$ \COMMENT{elementwise}
      \STATE \textbf{let} $U \Lambda U^T = \hat \Sigma$ \COMMENT{symmetric eigendecomposition}
      \STATE \textbf{sample} $X \in \R^{n \times d}$ iid unit Gaussians
      \STATE \textbf{return} $\sgn(U \Lambda_+^{1/2} X )$ \COMMENT{elementwise}
    \end{algorithmic}
    \end{algorithm}
\end{minipage}
\end{wrapfigure}

A natural way to represent a desired similarity matrix $\mat{M} \in \R^{n \times n}$ is to project it onto the set of representable matrices of binary vectors, which would correspond to a distribution one could sample from. Unfortunately, this approach is intractable as it would require solving a linear programming problem of size exponential in $n$. Instead, to approach the expressivity limits of binary HDC, we propose the following approach, given in Algorithm~\ref{alg:gaussianhv}. First, we sample $d$ independent multivariate Gaussians over $\R^n$;
% with covariance matrix $\hat\Sigma$; 
Our $n$ HDC vectors of length $d$ are then given by the signs of these Gaussians.
% a way to produce results that are close to $\mat{M}$ with an analogy of random fourier features, i.e. we sample basis hypervectors in the space according to some distributions such that
% \[
% \expect[\mat{M}^{\basis}] = \mat{M}\; .
% \]
The following lemma tells us how to make this produce a desired similarity matrix $\mat{M}$.
\begin{lemma}
Suppose $X, Y$ are jointly Gaussian zero-mean unit-variance random variables, then $\expect[\sgn(X)\sgn(Y)] = \frac{2}{\pi} \arcsin\left( \expect[X Y ] \right).$
% \[
% \textstyle
% \expect[\sgn(X)\sgn(Y)] = \frac{2}{\pi} \arcsin\left( \expect[X Y ] \right).
% \]
% \begingroup
%     \setlength\abovedisplayskip{2pt}
%     \setlength\belowdisplayskip{2pt}
%     \[
% \textstyle
% \expect[\sgn(X)\sgn(Y)] = \frac{2}{\pi} \arcsin\left( \expect[X Y ] \right).
% \]
% \endgroup

% and with covariance $\expect{X Y} = $
% Assume $\vec{x}=(x_1,\ldots, x_n)^\top$ with each element drawn from $\mathcal{N}(0,1)$, let $u=\vec{a}^\top\vec{x}, v=\vec{b}^\top\vec{x}$, then 
% \[
% \expect[\sgn(u)\sgn(v)] = 1-\frac{2}{\pi}\arccos(\theta),
% \]
% where $\theta=\arccos(\frac{\vec{a}^\top\vec{b}}{\|\vec{a}\|\cdot\|\vec{b}\|})$.
\end{lemma}
% \begin{proof}
% Note that 
% \[
% \begin{split}
%     &\expect[\sgn(u)\sgn(v)] = 1-\frac{\pi}{2}\arccos(\rho)\\
%     =&\prob(u>0,v>0)-\prob(u>0,v<0)\\
%     &+\prob(u<0,v<0)-\prob(u<0,v>0) \\
%     =&\prob(u>0,v>0)-(\prob(u>0) - \prob(u>0,v>0))\\
%     &+\prob(u<0,v<0)-(\prob(u<0)-\prob(u<0,v>0)) \\
%     =&2\prob(u>0,v>0)+2\prob(u<0,v<0)-\prob(u>0)\\
%     &-\prob(u<0)\\
%     =&2(\prob(u>0,v>0)+\prob(u<0,v<0))-1\\
%     =&2(\frac{\pi-\theta}{2\pi}+\frac{\pi-\theta}{2\pi})-1\\
%     =&1-\frac{2\theta}{\pi}\;.
% \end{split}
% \]
% \end{proof}
% \begin{proof}
% Without loss of generality let $U \sim \mathcal{N}(0,I)$ be a standard Gaussian over $\R^2$, and suppose that $X = a^T U$, $Y = b^T U$ for some vectors $a,b \in \R^2$ with $\|a\|=\|b\|=1$ and $a^T b = \expect[XY]$. Here, a geometric argument shows that $\prob(X \ge 0 \land Y \le 0) = \prob(a^T U \ge 0 \land b^T U \le 0) = \theta/(2 \pi)$, where $\theta$ is the angle between $a$ and $b$. An analogous analysis of the other three cases, combined with some straightforward trigonometry, proves the lemma.
% \end{proof}

From this lemma, it immediately follows that if the elementwise $\sin$ of $\frac{\pi}{2}\mat{M}$
% $\pi/2$ times the desired similarity matrix $\mat{M}$ 
is positive semi-definite, then Algorithm~\ref{alg:gaussianhv} produces hypervectors that, in expectation, exactly achieve $\mat{M}$; otherwise, some approximation to $\mat{M}$ is produced. It also immediately follows that Algorithm~\ref{alg:gaussianhv} can achieve more similarity matrices than the classical procedure of Lemma~\ref{lemma:classichdc}: while that lemma shows that the similarity matrix $\frac{4}{3} I_3 - \frac{1}{3} \mathbf{1}_{3 \times 3}$ cannot be achieved in expectation by classical HDC initialization, Algorithm~\ref{alg:gaussianhv} can achieve it as $\sin(\frac{\pi}{2} \cdot \frac{-1}{3}) = \frac{-1}{2}$ and $\frac{3}{2} I_3 - \frac{1}{2} \mathbf{1}_{3 \times 3}$ is positive semidefinite.

% \begin{theorem}
% Algorithm 1 satisfies that \tao{insert the algorithm here}.
% \[
% \expect[\text{sgn}(v_i)\text{sgn}(v_j)] = \mat{M}_{ij}
% \]
% \end{theorem}

Algorithm~\ref{alg:gaussianhv} gives us more freedom to achieve a wider range of similarity matrices; however, it does not tell us \emph{which} similarity matrix $\mat{M}$ to choose for a given task and whether $\sin(\frac{\pi}{2} \mat{M})$ is positive semi-definite or not. In this paper, we use the well-known RBF kernel \citep{vert2004rbf} to choose the similarity matrix between entities, but any similarity matrix appropriate for a task is applicable. %\tao{here needs some polish of the explanation, maybe connection to kernel methods, Chris?} 

% \tao{Some empirical results here, illustration figure, correction. if short of space}

%% file: sections/groupvsa.tex
\section{Group VSA}
\label{sec:groupvsa}
% \tao{need a table listing existing work including binary HDC, unit cycle VSA and our proposed VSA, together with the specific binding, bundling ex. operations}
So far we have shown how replacing existing initialization methods can approach the limits of binary HDC. However, as Lemma~\ref{lem:sim_hdc_fail} shows, binary HDC itself has inherent limits. Other known VSAs, such as the unit cycle VSA \citep{plate1994complexHDC}---in which the elements are complex numbers of absolute value $1$---can surpass these limits. However, this comes with the problem of a continuous space---requiring both approximation and significant hardware complexity overhead compared to binary HDC. 
In this section, we propose a new class of VSA, \emph{finite group VSA}, which effectively ``interpolates'' between them so as to bypass the similarity-representation limits of binary HDC without the need for a continuous space.
% It's common to increase the dimensionality of hypervectors until a satisfactory performance is achieved. However, as shown in the toy example, for HDC, without considering the similarity matrix of the basis, in some cases, it's not necessary to arrive at better performance by simply increasing the dimensionality. 

We start by defining a VSA, and then propose to use group structures for the elements of hypervectors as a different approach to improve the expressivity of VSA. Binary HDC can be considered as a special case of our construction corresponding to the 2-element group. %Furthermore, we also provide an axiomatic approach to use group VSA. 

% There are 4 things we need to characterize for group VSA, namely, VSA element, binding, bundling and similarity measure. 

\begin{definition}
\label{def:vsa}
A group VSA is a tuple $(G, \mu, \otimes, \mathcal{S}, \oplus)$, where $G$ is some measurable set of symbols, $\otimes: G \times G \rightarrow G$ is the binding operator, $\mathcal{S}: G \times G \rightarrow \R$ is a symmetric similarity operator, and $\oplus: G^{< \omega} \rightarrow G$ is the bundling operator (which maps a finite sequence of symbols to a symbol). A VSA must have the following properties:
\begin{itemize}[nosep, leftmargin=*]
    \item \textbf{Binding.} $(G, \otimes)$ is a group,\! i.e., binding is associative over $G$ and has inverse and identity elements.
    \item \textbf{Similarity to self.} $\simi(x,x) = 1$ for all $x \in G$.
    \item \textbf{Similarity preserved by binding.} $\simi(g \otimes x,g \otimes y) = \simi(x \otimes g,y \otimes g) = \simi(x,y)$ for all $g,x,y \in G$.
    \item \textbf{Similarity extensible to an inner product.} There exists a finite-dimensional Hilbert space $V$ over $\R$ and an embedding $\psi: G \rightarrow V$ such that $\simi(x,y) = \langle \psi(x), \psi(y) \rangle$ for all $x,y \in G$. Equivalently, $\sum_{i=1}^n \sum_{j=1}^n c_i c_j \simi(x_i, x_j) \ge 0$ for any $x_1, \ldots, x_n \in G$ and scalars $c_1, \ldots, c_n \in \R$.
    % \begingroup
    % \setlength\abovedisplayskip{2pt}
    % \setlength\belowdisplayskip{2pt}
    % \[
    %     \textstyle
    %     \sum_{i=1}^n \sum_{j=1}^n c_i c_j \simi(x_i, x_j) \ge 0.
    % \]
    % \endgroup
    \item \textbf{Random vectors are dissimilar to any other vector.} $\expect_{g \sim \operatorname{Uniform}(G)}[\simi(g,x)] = 0$ for any $x \in G$. 
    \item \textbf{Bundling.} Bundling of $x_1, \ldots, x_m \in G$ returns $\bigoplus_{i=1}^m x_i =\arg \max_{g \in G} \;
        \sum_{i=1}^m  \simi( g , x_i ) \label{eqn:bundling}$
    % \begingroup
    % \setlength\abovedisplayskip{2pt}
    % \setlength\belowdisplayskip{2pt}
    % \begin{equation}
    %     \textstyle
    %     \bigoplus_{i=1}^m x_i
    %     =
    %     \arg \max_{g \in G} \;
    %     \sum_{i=1}^m  \simi( g , x_i ) \label{eqn:bundling}
    % \end{equation}
    % \endgroup
    or one of the maxima in case of a tie.
\end{itemize}
\end{definition}

To compute using a VSA of dimension $D$, we use hypervectors in $G^D$, extend binding and bundling to act elementwise on these hypervectors, and extend similarity to compute the average similarity over the dimensions as $\mathcal{S}([x_1, \ldots, x_D],[y_1, \ldots, y_D]) = \frac{1}{D} \sum_{i=1}^D \mathcal{S}(x_i, y_i)$.
% \begingroup
% \setlength\abovedisplayskip{2pt}
% \setlength\belowdisplayskip{2pt}
% \[
%     \textstyle
%     \mathcal{S}([x_1, \ldots, x_D],[y_1, \ldots, y_D]) = \frac{1}{D} \sum_{i=1}^D \mathcal{S}(x_i, y_i).
% \]
% \endgroup
It is easy to see that binary HDC is equivalent to a VSA where $G = \{-1,1\}$, $\otimes$ is multiplication, and $\mathcal{S}(x,y) = xy$.
Similarly, a unit-cycle VSA (i.e., FHRR) has $G = \{ z \in \mathbb{C} \mid |z| = 1 \}$, $\otimes$ as complex multiplication, and $\mathcal{S}(x,y) = \operatorname{Re}(x^* y)$.
Most other schemes called ``VSAs'' in literature fall under our definition, e.g., \citet{Gayler1998MultiplicativeBR}, with few exceptions \citep{plate1995holographic} that violate the group requirement.
In order to run efficiently on hardware, we add the following restriction.

\begin{definition}
A finite group VSA is a VSA where $G$ is finite. That is, $(G, \otimes)$ is a finite group.
\end{definition}

On hardware, finite $G$ allows the VSA elements to be represented exactly and lets the VSA operations be computed exactly. The hardware cost will depend on the size of $G$.
% It is also desirable---although as we will see not always achievable---for the codomain of $\mathcal{S}$ to be the rational numbers, as then similarities can be computed exactly using rational-number arithmetic.
In many cases, we would like binding to preserve similarity in a stronger sense than that guaranteed by Definition~\ref{def:vsa}.
It is often desirable that the similarity after binding is the product of the similarities before binding, i.e.,
% $\mathcal{S}(x_1 \otimes  x_2, y_1 \otimes  y_2) = \mathcal{S}(x_1, y_1) \cdot \mathcal{S}(x_2, y_2)$,
\begingroup
\setlength\abovedisplayskip{2pt}
\setlength\belowdisplayskip{2pt}
\begin{equation}
    \mathcal{S}(x_1 \otimes  x_2, y_1 \otimes  y_2) = \mathcal{S}(x_1, y_1) \cdot \mathcal{S}(x_2, y_2); \label{eqn:productsimple}
\end{equation}
\endgroup
% For example, this is desirable when binding hypervectors that originate from different features \citesomething{}. 
this would make $\otimes$ behave like a tensor product space with inner product given by $\simi$. This property is particularly important, usually when an object consists of multiple features, we will bind these features so as to derive a representative vector for the object, this property ensures that two objects with multiple pairs of similar features to have similar representative vectors. 

Of course this equation is \emph{not} guaranteed to hold for elements of $G$ in general (e.g., when $x_1 \otimes  x_2 = y_1 \otimes  y_2$); however, most VSAs can approximate this behavior by adding an extra randomization step.

% \begin{definition}
% A whitening transformation \fixme{think of a better name?} of a VSA is a distribution $A$ over automorphisms of $(G, \otimes, \mathcal{S})$ such that for any $x_1, x_2, y_1, y_2 \in G$,
% \begin{align*}
%     &\textstyle
%     \expect_{\alpha \sim A}[\mathcal{S}(x_1  \otimes \alpha(x_2), y_1  \otimes \alpha(y_2))]
%     \\&\textstyle\hspace{2em}=
%     \expect_{\alpha \sim A}[\mathcal{S}(\alpha(x_1) \otimes  x_2, \alpha(y_1)  \otimes y_2)]
%     \\&\textstyle\hspace{2em}=
%     \mathcal{S}(x_1, y_1) \cdot \mathcal{S}(x_2, y_2).
% \end{align*}
% \end{definition}
\begin{definition}
Let $A$ denote the uniform distribution over automorphisms of $(G, \otimes, \mathcal{S})$. 
Then we say that VSA has the \emph{product property} if for any $x_1, x_2, y_1, y_2 \in G$,
% \[
% \textstyle
% \expect_{\alpha \sim A}[\mathcal{S}(x_1  \otimes \alpha(x_2), y_1 \otimes \alpha(y_2))] =
%     \expect_{\alpha \sim A}[\mathcal{S}(\alpha(x_1) \otimes  x_2, \alpha(y_1)  \otimes y_2)]= \mathcal{S}(x_1, y_1) \cdot \mathcal{S}(x_2, y_2).
% \]
\begingroup
    \setlength\abovedisplayskip{2pt}
    \setlength\belowdisplayskip{2pt}
    \[
\textstyle
\expect_{\alpha \sim A}[\mathcal{S}(x_1  \otimes \alpha(x_2), y_1 \otimes \alpha(y_2))] =
    \expect_{\alpha \sim A}[\mathcal{S}(\alpha(x_1) \otimes  x_2, \alpha(y_1)  \otimes y_2)]= \mathcal{S}(x_1, y_1) \cdot \mathcal{S}(x_2, y_2).
\]
\endgroup
% \begin{align*}
%     &\textstyle
%     \expect_{\alpha \sim A}[\mathcal{S}(x_1  \otimes \alpha(x_2), y_1  \otimes \alpha(y_2))]
%     \\&\textstyle\hspace{2em}=
%     \expect_{\alpha \sim A}[\mathcal{S}(\alpha(x_1) \otimes  x_2, \alpha(y_1)  \otimes y_2)]
%     \\&\textstyle\hspace{2em}=
%     \mathcal{S}(x_1, y_1) \cdot \mathcal{S}(x_2, y_2).
% \end{align*}
\end{definition}
% It is easy to see that for binary HDC we can choose $A$ to be the distribution supported on just the identity automorphism (the only automorphism of $\mathbb{Z}/2\mathbb{Z}$), while for the unit cycle VSA we can choose $A$ to assign probability $1/2$ to the identity automorphism ($z \mapsto z$) and probability $1/2$ to the complex conjugate automorphism ($z \mapsto z^*$).
It is easy to see that this holds for binary HDC, as the identity map is the only automorphism of $\mathbb{Z}/2\mathbb{Z}$ and (\ref{eqn:productsimple}) holds for binary HDC; this also holds for the unit cycle VSA, where the only automorphisms are the identity ($z \mapsto z$) and the complex conjugate automorphism ($z \mapsto z^*$).
Rather than using this transformation directly, if one exists we can ensure this ``product property'' holds by initializing our hypervectors appropriately: if we sample hypervectors $(x_1, y_1)$ at random with independent entries and independently of $(x_2, y_2)$ and both distributions are invariant under automorphisms, then $\expect[\mathcal{S}(x_1 \otimes  x_2, y_1 \otimes  y_2)] = \expect[\mathcal{S}(x_1, y_1)] \cdot \expect[\mathcal{S}(x_2, y_2)]$.
% \[
%     \textstyle
%     \expect[\mathcal{S}(x_1 \otimes  x_2, y_1 \otimes  y_2)] = \expect[\mathcal{S}(x_1, y_1)] \cdot \expect[\mathcal{S}(x_2, y_2)].
% \]
% Similar to the HDC case, we give the formal definition of expressivity for general VSAs here:
% \begin{definition}
% \label{def:vsa-expressivity}
% The expressivity $\mathcal{E}(\mathcal{V})$ of a set of VSAs $\mathcal{V}$ is defined as the set of similarity matrices that can be approximated arbitrarily well by some hypervectors in the VSA. Explicitly, for a $n \times n$ matrix $\mat{M}$, we say $\mat{M} \in \mathcal{E}(\mathcal{V})$ iff for every $\epsilon > 0$, there exists a VSA in $\mathcal{V}$, a size $d$, and $n$ length-$d$ vectors $x_1, x_2, \ldots, x_n$ in that VSA such that for all $i$ and $j$, $| \mat{M}_{ij} - \simi(x_i, x_j)| \le \epsilon$. 
% \end{definition}

\subsection{Constructing a VSA from a finite group}

At first glance, the definition of a group VSA may seem open-ended, offering little guidance as to what the limitations of finite group VSAs may be and how they can be constructed. Surprisingly, we can fully characterize the finite group VSAs through \emph{representation theory}. We start by introducing definitions specialized to finite-dimensional complex representations, before stating the full theorem.

\begin{definition}[\citet{james2001representations,fulton2013representation}]
A \emph{representation} of a group $G$ over $\mathbb{C}^n$ is a group homomorphism $\rho$ from $G$ to $\mathbb{C}^{n \times n}$
% i.e. a function $\rho: G \rightarrow \mathbb{C}^{n \times n}$ 
such that $\rho(gh) = \rho(g) \rho(h)$ for all $g,h \in G$.

The \emph{character} of a representation $\rho$ is the function $\chi: G \rightarrow \mathbb{C}$ given by $\chi(g) = \operatorname{Tr}(\rho(g))$. The representation (and corresponding character) is said to be \emph{irreducible} if no proper subspace of $\mathbb{C}^n$ is preserved by the group action. The \emph{trivial representation}, had by all groups, is $\rho: G \rightarrow \mathbb{C}^{1 \times 1}$ with $\rho(g) = 1$ and $\chi(g) = 1$.
\end{definition}

It is a standard result that each finite group possesses a finite number of irreducible characters equal to the number of conjugacy classes of the group \citep{serre1977linear,fulton2013representation}.

\begin{theorem}
\label{thm:vsacharacter}
Let $(G, \otimes)$ be a finite group, and let $X$ denote the set of its non-trivial irreducible characters.
Let $\alpha\!\!:\!\! X\!\!\!\rightarrow\!\! \R_{\ge 0}$ be some function that assigns a non-negative weight to each of the characters. Then, if we set $\mathcal{S}$ as $\mathcal{S}(g, h)\! =\! \frac{\sum_{\chi \in X} \alpha(\chi) \cdot \operatorname{Re}(\chi(g^{-1} \otimes h))}{\sum_{\chi \in X} \alpha(\chi) \cdot \chi(\mathbf{1})}$,
% \[
%     \textstyle
%     \mathcal{S}(g, h) = \frac{\sum_{\chi \in X} \alpha(\chi) \cdot \operatorname{Re}(\chi(g^{-1} \otimes h))}{\sum_{\chi \in X} \alpha(\chi) \cdot \chi(\mathbf{1})},
% \]
% \begingroup
% \setlength\abovedisplayskip{2pt}
% \setlength\belowdisplayskip{2pt}
% \[
%     \textstyle
%     \mathcal{S}(g, h) = \frac{\sum_{\chi \in X} \alpha(\chi) \cdot \operatorname{Re}(\chi(g^{-1} \otimes h))}{\sum_{\chi \in X} \alpha(\chi) \cdot \chi(\mathbf{1})},
% \]
% \endgroup
where the inverse and unit $\mathbf{1}$ are those of the group,
% $(G, \otimes)$, 
and define bundling $\oplus$ as given in (\ref{eqn:bundling}), then $(G, \otimes, \mathcal{S}, \oplus)$ is a finite group VSA. Any finite group VSA can be constructed in this way. If in this construction $\alpha$ is supported on only one character $\chi$, i.e. $\mathcal{S}(g, h) =\operatorname{Re}(\chi(g^{-1} \otimes h))/ \chi(\mathbf{1})$, then the VSA will have the product property.
\end{theorem}

This construction makes it seem as though finite-group VSAs with the product property may be a restricted subset, which could be less expressive. The following result shows that this is not the case.
\begin{statement}
Let $\mat{M}$ be a similarity matrix expressible by a finite group VSA. Then there exists a finite group VSA that has the product property and can also express $\mat{M}$.
\end{statement}
% \begin{proof}
% Suppose the first VSA's group is $G$ and has irreducible characters $\chi_1, \chi_2, \ldots, \chi_k$. Then the group $G^k$ consisting of the direct product of $k$ copies of the group $G$, together with a similarity function $\simi((x_1, \ldots, x_k), (y_1, \ldots, y_k)) \propto \prod_{i=1}^k \chi_i(x_i^{-1} \otimes y_i)$ will both have the product property (as its similarity matrix is proportional to a single character) and can express any similarity matrix the first VSA can.
% \end{proof}

% \textbf{\textcolor{red}{[cmd353 finished pass here]}}

\subsection{Cyclic Group VSA}
Most of our work with group VSAs in this paper will use the cyclic group $G=\mathbb{Z}/n\mathbb{Z}=\{0, 1, \cdots, n-1\}$, as Definition \ref{def:vsa} indicates, we first provide an embedding $\psi: G \rightarrow V$ to a finite-dimensional Hilbert space $V$ over $\R$.
Let $\psi(x) = \left(\cos( 2\pi x / n ), \sin(2\pi x / n) \right)$.
% \begin{definition}
% Consider $\mathcal{D}$ as the unit cycle in $\R^2$ and $\psi: G \rightarrow \mathcal{D}$ defined as 
% \[
% \psi(x) = (\cos\frac{2\pi x}{n}, \sin\frac{2\pi x}{n}).
% \]
% Further more, we consider the inverse map $\psi^{-1}$ as a mapping $\psi^{-1}: \mathcal{D} \rightarrow [0, n)$.
% % Since there is a bijection, we denote the inverse map as $\psi^{-1}: V \rightarrow G$.
% \end{definition}
\begin{definition} \label{def:cyclicvsa}
The standard cyclic group VSA is given by:
\begin{itemize}[nosep, leftmargin=*]
    \item The symbol set $G=\mathbb{Z}/n\mathbb{Z}=\{0, 1, \cdots, n\!-\!1\}$ with addition modulo $n$ as binding operation $\otimes$.
    \item Similarity is defined as 
    $\simi(x,y) = \langle \psi(x), \psi(y) \rangle = \cos(2\pi(x-y)/n).$
    % \item Following Eq. \ref{eqn:bundling}, bundling can be derived as 
    % \[
    %     \textstyle
    %     \bigoplus_{i=1}^m x_i
    %     = \floor*{\psi^{-1}(\frac{\sum_{i=1}^m  \psi(x_i)}{\|\sum_{i=1}^m  \psi(x_i)\|})},
    % \]
    % which is essentially a projection of $\sum_{i=1}^m  \psi(x_i)$ to $\psi(G)$.
\end{itemize}
\end{definition}
This cyclic group VSA is in some sense a ``subset'' of the unit cycle VSA, and as $n$ goes to infinity, it approximates the the unit cycle VSA arbitrarily well \citep{plate1994complexHDC}, serving as an interpolation between the binary HDC and the unit cycle VSA.
As a straightforward consequence, any $\mat{M}$ that can be expressed by this VSA can also be expressed by the unit cycle VSA.
To compute with this VSA, we follow the procedure in Section \ref{sec:groupvsa}: use hypervectors in $G^D$, and extend similarity, binding and bundling operations accordingly. 
Similar to the HDC case, we utilize random Fourier features for a better basis hypervectors initialization with minor modifications of Algorithm \ref{alg:gaussianhv}; we replace the $\sgn$ function in the last step with the ($n$th) quantile function of a Gaussian so as to map into $G$. 

Note that as Theorem~\ref{thm:vsacharacter} shows, this setup is not the only VSA over the cyclic group. Indeed, for any distribution $\alpha$ over $\{1,\ldots,n-1\}$, the similarity function $\simi(x,y) = \sum_{k=1}^{n-1} \alpha(k) \cos\left(\frac{2\pi(x-y) k}{n} \right)$
% \[
%     \textstyle
%     \simi(x,y) = \sum_{k=1}^{n-1} \alpha(k) \cos\left(\frac{2\pi(x-y) k}{n} \right)
% \]
would yield a finite group VSA. We focus on the VSA of Definition~\ref{def:cyclicvsa} because it satisfies the product property, and all other VSAs on $G$ that do so are either isomorphic to it or isomorphic to the same construction with a smaller $n$.

\subsection{Non-Abelian Finite Group VSAs}
Our analysis of cyclic group VSAs from the previous section extends naturally to cover all finite Abelian groups (i.e. groups in which $\otimes$ is commutative), since it is a classic result that every finite Abelian group factors as the direct product of cyclic groups. It is natural to ask: what about non-Abelian groups? Because they simplify both representation and computation, it would be convenient if we could restrict our attention to Abelian groups only. Unfortunately, the following two statements together show that non-Abelian groups can be strictly more expressive than Abelian groups.

\begin{statement}
\label{stmt:abelian}
Any similarity matrix $\mat{M}$ that can be expressed by a finite Abelian group VSA can be expressed by the unit-cycle VSA ($G = \{z \in C \mid |z| = 1\}$, $x \otimes y = xy$, $\simi(x,y) = \operatorname{Re}(x^* y)$).
\end{statement}

\begin{statement}
\label{stmt:nonabelian}
There exists a similarity matrix $\mat{M}$ that can be expressed by a VSA over the (non-Abelian) binary icosahedral group, but not by the unit-cycle VSA (i.e., FHRR).
\end{statement}

Statement~\ref{stmt:abelian} follows from the standard representation-theoretic result that all irreducible representations of a finite Abelian group are one-dimensional, while Statement~\ref{stmt:nonabelian} is proved by direct construction. While these results do show that, non-Abelian finite group VSAs are ``more powerful'' than Abelian finite group VSAs, the additional complexity needed to unlock this power seems not worthwhile for our applications, where unit-cycle VSA already performs well---so, in our experiments we focus solely on the cyclic group.
We leave exploration of non-Abelian VSAs to future work.

%% file: sections/sgd.tex
\section{Learning via SGD Instead of Bundling}
\label{sec:sgd}

Prior works train an HDC model via bundling hypervectors in the same class $\mathbb{T}_{c} = \{ t_{i} | \text{label}(t_{i})=c \}$.
This is based on the fundamental assumption about bundling that the class representative $s_c$ is similar to each $t_i$.
We find that it is \emph{not always true}, depending on the number of vectors being bundled. 

% Let $2k+1$ be the number of hypervectors in the set $\mathbb{T}_c$. 
% The odd number of vectors avoids breaking a tie. 
% Assume these hypervectors are nearly orthogonal, e.g., similar to randomly drawn from a hyperspace. 
Suppose a set $\mathbb{T}_c$ has $2k+1$ (avoids a tie) unrelated hypervectors,
we can theoretically calculate\footnote{The calculation is in appendix.} the expected angle $\theta$ between $s_c$ and a randomly selected hypervector $t_{i}$:
$ \theta_{2k+1} = \arccos\left(\binom{2k}{k}/ 2^{2k} \right)$.
This indicates that the class vector learned from bundling will be nearly orthogonal to each hypervector in the class and no longer be its representative as we increase $k$.

As an alternative, we propose to leverage stochastic gradient decent (SGD) to learn a linear classifier (same precision).
% , which is a linear model (same precision) with the number of weights being $\# \operatorname{class} \times D$.
Take binary HDC as an example, the classifier is a binarized matrix multiplication at inference time, i.e., $O = X \cdot\sgn(W)$, where $X$ is the binary hypervector and $W$ is the weight matrix. 
During the back propagation, we use the straight-through estimator~\citep{hubara2016bnn} to approximate the gradient of the sign function: 
$\partial \sgn(W)/\partial W := 1~\textit{if}~\left | W \right | < 1~\textit{else}~0$.

The inference cost of an HDC model remains the same as the bundling paradigm since they are both binary. The model is still trained for one or few epochs so the SGD approach incurs minor training overhead. We defer the SGD learning process of group VSAs to Appendix. 

%% file: sections/experiments.tex
\section{Experiments}
\begin{table*}[tb]
\centering
% \addtolength{\tabcolsep}{-4.5pt}
\caption{Comparison on test accuracy of proposed methods to SOTA HDC$^\dagger$ \citep{imani2019hdcframework}, dynamic HDC$^*$ \citep{chuang2020dynamicHDC} and 1-bit RFF perceptron. Dimension of hypervectors is 10,000. 1-Epo: 1-Epoch, 10-Epo: 10-Epoch.}
\label{tab:model-accuracy}
\scalebox{1.0}{\begin{tabular}{lcccccccc}
\toprule
\textbf{Dataset}&\multicolumn{2}{c}{ISOLET}&\multicolumn{2}{c}{UCIHAR}&\multicolumn{2}{c}{MNIST}&\multicolumn{2}{c}{Fashion-MNIST}\\
\cmidrule(l){2-3}
\cmidrule(l){4-5}
\cmidrule(l){6-7}
\cmidrule(l){8-9}
{\textbf{Acc}(\%)} & \textbf{1-Epo}& \textbf{10-Epo}& \textbf{1-Epo}& \textbf{10-Epo}& \textbf{1-Epo}& \textbf{10-Epo}&\textbf{1-Epo}& \textbf{10-Epo}\\
\midrule
 Percep. & 82.8 & 90.1 & 69.3 & 91.4 & 94.3
& 94.3 & 79.5 & 79.5 \\
 HDC$^\dagger$ & 85.6 & 91.5 & 87.3
& 95.7 & NA & 89.0 & NA & NA\\
 RFF HDC  & 90.6 &  94.4 &  93.8 & 95.7  & 95.4 & 95.4 & 83.4& 84.0  \\
 RFF G($2^3$)-VSA & 93.1 & 94.4  & 95.1 & 95.6 & 96.3 
& 95.7 & 85.4 & \textbf{86.7} \\
 RFF G($2^4$)-VSA & \textbf{94.4} & \textbf{96.0}  & \textbf{95.5} & \textbf{96.6} & \textbf{96.5}
& \textbf{96.6} & \textbf{87.4} & 86.5 \\
\bottomrule
\end{tabular}}
\vspace{-1em}
\end{table*}
% \tao{do we still need a figure illustrating the group vsa performance versus the order of group as some tradeoff curves.}
% 1. Datasets --- MNIST, Fashion-MNIST, ISOLET, UCIHAR, and potentially a large scale dataset, FACE

% 2. Compare HDC, group HDC, perceptron on binary hign-dimensional RFFs, kernal perceptron, vanilla perceptron on raw data.

% 3. Compare the number of xnors on the critical path (circuit depth).

% \subsection{Model Performance}
\textbf{Datasets.}
We evaluate the performance of proposed methods on two conventional HDC datasets, ISOLET~\citep{Dua2019isolet} and UCIHAR~\citep{anguita2012ucihar}.
We also evaluate our method on MNIST and Fashion-MNIST~\citep{xiao2017fashionmnist}, which are more challenging for HDC. 
ISOLET is a speech recognition dataset where each sample is an audio signal with 617 features. Each feature is in the range of $[-1, 1]$. The dataset has 7719 samples in total. The goal is predicting which letter-name was spoken.
UCIHAR is a human activity recognition database, each sample of which contains 561 features collected from smartphone sensors. The features are also in the range of $[-1, 1]$. The database has 10299 samples. The task is predicting which type of activity a human was performing.

\textbf{Setups.}
For ISOLET and UCIHAR, we quantize the features to 8 bits before encoding. 
We initialize a 10,000-dimensional basis hypervector for each $\{0,\cdots, 255\}$ feature value, then encode raw inputs as described in Section~\ref{sec:rff} or \ref{sec:groupvsa}. 
During the training stage, we use a learning rate of 0.01 and train classifiers for 10 epochs. 
% More details on hyper-parameters are provided in Appendix. 
We compare RFF-HDC and group VSA of order $2^3$ and $2^4$ with SOTA HDC \citep{imani2019hdcframework, chuang2020dynamicHDC} \footnote{\citet{alejandro2021onlineHD} seem to have a better result on MNIST in Figure 7, but no concrete numbers are reported.} that propose iteratively updating the class vectors through misclassified examples.
We also compare HDC to a perceptron \citep{rosenblatt1958perceptron} where inputs are 10,000-dimensional binary RFFs generated from raw data. 
We train on Intel Xeon CPUs.

\textbf{Results.} 1- and 10-epoch test accuracies are in Table \ref{tab:model-accuracy}, which yield three key observations:
\begin{tightlist}{1em}{0.3em}{0.3em}
    \item \textbf{RFF HDC already improves non-trivially over the baseline SOTA HDC.}
With basis hypervectors initialized from the similarity matrix constructed from pixel similarities, our method improves the MNIST model accuracy by $6.4\%$ compared to the SOTA. It also for the first time enables HDC learning on Fashion-MNIST, a more challenging task, obtaining 84\% final accuracy.
    \item \textbf{Group VSA improves the model accuracy further.} By extending HDC to group VSA, the vector elements are in a higher complexity so that it can more precisely approximate the target similarity matrix than binary hypervectors. Figure~\ref{fig:tradeoff} shows that when there are 8 or 16 elements in the group, meaning the precision of each element in the hypervector is 3 or 4 bits, the proposed group VSA strikes a good trade-off between accuracy and complexity. It can further outperform RFF HDC by at least 1\% across various datasets.
    \item \textbf{Our HDC models learned from a single pass over the data achieve high accuracy.}
In all the evaluated tasks, our proposed RFF HDC or the extended group VSA can both achieve approximately the final accuracy in one single epoch. 
In some cases, e.g., Fashion-MNIST, the group VSA can even obtain a better quality with one single pass.
The single-pass model accuracy of HDC is significantly better than the baseline perceptrons, especially on the ISOLET and UCIHAR datasets, which has an at least 8\% gap.
This evidence shows that HDC learning has an impressive data efficiency.
This capability of single-pass learning is consistent with finding in prior works \citep{alejandro2021onlineHD, imani2019hdcframework}.
\end{tightlist}

% \tao{cifar10 results}

% \begin{table}[htb]
%     % \renewcommand{\arraystretch}{1.2}
%     \addtolength{\tabcolsep}{-4.5pt}
%     \caption{Comparison on test accuracy of proposed methods on CIFAR10 dataset. Dimension of hypervectors is 10,000. 1E-Acc: 1-Epoch Accuracy, F-Acc: 10-Epoch Accuracy.}
%     \label{tab:model-accuracy}
%     \centering
%     \begin{tabular}{ccc}
%     \toprule
%     \textbf{Method} & \textbf{1E-Acc}(\%) & \textbf{F-Acc}(\%) \\
%     \midrule
%     RFF HDC &  &  \\
%     RFF G($2^3$)-VSA &  &  \\
%     RFF G($2^4$)-VSA & \textbf{} & \textbf{} \\
%     \bottomrule
%     \end{tabular}
% \end{table}

\subsection{Circuit-Depth Complexity}
\begin{table}[t]
% \vspace{-3em}
\begin{minipage}[c]{0.49\textwidth} 
    \footnotesize
    \addtolength{\tabcolsep}{-9pt}
    \caption{Analysis of circuit-depth complexity of binary HDC and 1-bit RFF perceptron.}
    \label{tab:cdc}
    \begin{tabular}{cc}
    \toprule
    Method & CDC \\
    \midrule
    Percep. & $91 + 96 \cdot \log_{2}N + \frac{3}{2}\log_{2}D \cdot (1+\log_{2}D)$  \\
    \midrule
    HDC & $\log_{2}N + 1 + \frac{3}{2}\log_{2}D \cdot (1+\log_{2}D)$ \\
    \midrule
    G($2^n$)-VSA & $3n\log_{2}N + 24\log_{2}D$ \\
    \bottomrule
    \end{tabular}
\end{minipage}% 
    \hfill
    % \hspace{-0.38in}
\begin{minipage}[c]{0.5\textwidth} 
    % \vspace{-2.2em}
    \begin{figure}[H]
    \centering
    \includegraphics[width=\columnwidth]{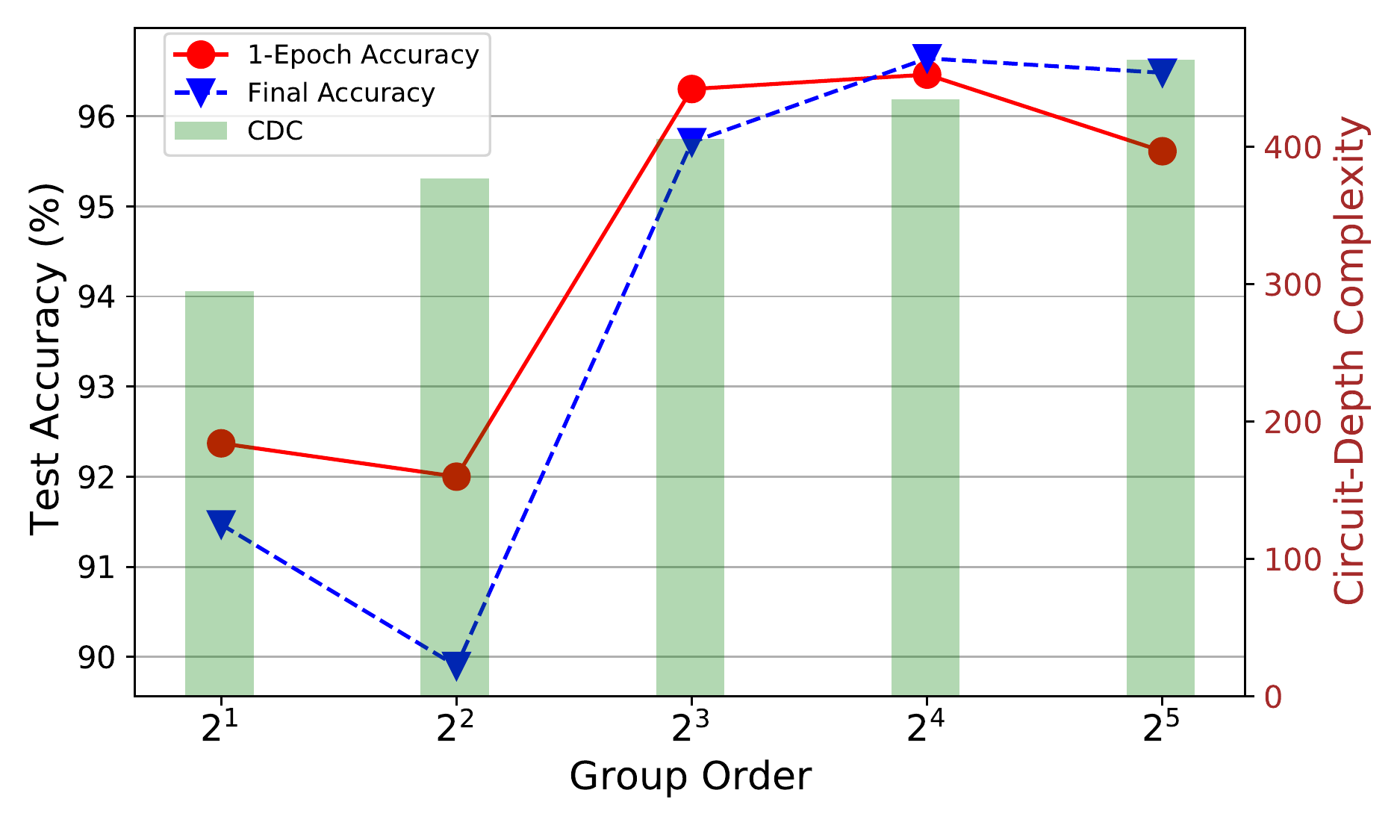}
    \vspace{-2em}
    \caption{Cyclic Group VSA on MNIST.}
    \label{fig:tradeoff}
    \end{figure}
\end{minipage}
\vspace{-3em}
\end{table}
% as a subsection in experiment and show this performance together in result table
% 1. How does HDC/group HDC execute on hardware
% 2. Explain why HDC has ultra low latency (energy cost?) during inference.
% There is a well defined circuit-depth complexity in theoretical computer science
%\zz{is there a HW paper we can cite which demonstrates the speed advantage of HDC? if so, perhaps we can quote some numbers at the end of the bin HDC paragraph to further justify our analysis.} 
% \zz{some reviewers may ask about the cost comparison (area/power). shall we clarify why we only focus on the circuit depth?}
To quantify the potential hardware latency of HDC,
we analyze its \emph{circuit-depth complexity} (CDC) in Table~\ref{tab:cdc},
defined as the length of the longest path from the input to the output (measured by the number of two-input gates along the path).
CDC is commonly used to analyze the complexity of Boolean functions.
We further assume that operations without data dependencies are all in parallel.
% This makes CDC independent of hardware design choices such as tiling.
Let $N$ be the feature vector length, 
e.g., $N=784$ for an MNIST image, 
$D$ be the hypervector dimension. 

\textbf{Binary HDC.} The encoding stage binds all feature hypervectors, which can be implemented in a tree structure.
The depth of a single binding operation (XNOR) is $1$.
The total depth is therefore $\log_{2}N$. 
Computing the similarity includes a binding and a bit counting.
% Computing the similarity between an encoded hypervector and a class vector includes one binding operation and one population count. 
% citation of the popcount come from: https://en.wikipedia.org/wiki/Adder_(electronics)#cite_note-Adder-5
We assume a B-bit ripple carry adder, a chain of full adders, has a depth of $3 \cdot B$~\citep{satpathy2016adder}. 
Therefore a D-bit population count has a depth of 
$3 + 6 + \cdots + 3\log_{2}D = \frac{3}{2}\log_{2}D \cdot (1+ \log_{2}D)$.

% \begin{figure}[tb]
% \centering
% \includegraphics[width=\columnwidth]{figs/tradeoff.pdf}
% \vspace{-10mm}
% \caption{Cyclic Group VSA of different orders on MNIST.}
% \label{fig:tradeoff}
% \vspace{-5mm}
% \end{figure}

\textbf{Cyclic Group VSAs.} For a cyclic group VSA of order $2^n$, the depth of a single binding is $3n$ as it is an addition over the group. Therefore, binding all features has a depth of $3n\log_{2}N$. For similarity computations, we precompute the similarity matrix, which consists of $\simi(x,y), \forall x,y\in G$ in 8-bit numbers. Hence, the depth of computing the similarity for hypervectors is $3\cdot 8\cdot\log_{2}D$.

\textbf{1-bit RFF perceptron.} Projecting a feature vector onto a selected basis requires a depth of a 32-bit multiplier and an adder. 
A 32-bit Wallece tree multiplier~\citep{wallace1964multiplier} has roughly a depth of 45.
A 32-bit ripple carry adder has a depth of 96. 
A cosine operation generating random fourier features requires about the same depth as a 32-bit multiplier if computing with the CORDIC algorithm~\citep{volder1959cordic}. Since the perceptron is 1-bit, computing the distance has the same depth as that of HDC.

As a result, binary HDC has a CDC of 295 on MNIST and the cyclic group G($2^3$) VSA has 405. The complexity of HDC is 4.4$\times$ lower than a 1-bit RFF perceptron with a depth of 1299, while CDC of the cyclic group G($2^3$) VSA is 3.2$\times$ lower. HDC and group VSA are much faster in potential. In Figure~\ref{fig:tradeoff}, we plot the performance and CDC of a cyclic group VSA when the order varies. Our code is available on github \footnote{\url{https://github.com/Cornell-RelaxML/Hyperdimensional-Computing}}.

\subsection{Discussion}
The performance of various HDC/VSA methods is closely related to the set of its expressible similarity matrices. As a matter of fact, the required similarity matrix to learn (some) tasks in the paper might already be covered by (or close to in terms of Frobenius norm) the set of expressible similarity matrices of the $10$k-dimensional RFF HDC. Hence, the improvement from group-VSA can be limited compared to RFF HDC. If instead considering a $1$k-dimensional binary RFF HDC with a smaller set of expressible similarity matrices, group-VSA demonstrates a much better accuracy improvement. For example, on MNIST, $1$k-dimensional binary RFF HDC achieves $65.59\%$ 10-epoch test accuracy on MNIST. G($2^3$)-VSA, meanwhile, achieves $88.61\%$, and G($2^4$)-VSA achieves $92.56\%$ test accuracy.

The circuit depth serves as a preliminary analysis on the hardware complexity of HDC. While other efficient circuits, e.g., a parallel adder instead of a ripple carry adder, will have lower depth and make HDC attractive further, we avoid being over-optimistic on the estimation. For a practical hardware implementation, better circuits should be applied.
Besides, circuit depth only reflects the latency. In the future, an estimation on the number of operations will reflect the energy or circuit area and will further improve the analysis.

%% file: sections/conclusion.tex
\section{Conclusion}

From our theoretical analysis, there is a clear connection between the class of expressible similarity matrices and the expressivity of HDC/VSA. This new notion of expressivity reveals the limits of HDC that computes with binary hypervectors, and meanwhile provides a hint on how we can improve it.
The nontrivial improvement from group VSA and the proposed techniques on HDC across various benchmarks 
% including MNIST and fashion-MNIST
% MNIST and the demonstrated feasibility on fashion-MNIST 
suggests that this notion paves a new way towards the future development of HDC/VSA.

%% file: sections/acknowledgement.tex
\begin{ack}
%Use unnumbered first level headings for the acknowledgments. All acknowledgments go at the end of the paper before the list of references. Moreover, you are required to declare funding (financial activities supporting the submitted work) and competing interests (related financial activities outside the submitted work). More information about this disclosure can be found at: \url{https://neurips.cc/Conferences/2022/PaperInformation/FundingDisclosure}.

%Do {\bf not} include this section in the anonymized submission, only in the final paper. You can use the \texttt{ack} environment provided in the style file to autmoatically hide this section in the anonymized submission.

This work is supported in part by NSF Awards IIS-2008102 and CCF-2007832, and by CRISP, one of six centers in JUMP, a Semiconductor Research Corporation program sponsored by DARPA. The authors would like to thank Denis Kleyko from the Redwood Center for Theoretical Neuroscience at UC Berkeley and researchers from VSAONLINE for providing valuable feedbacks on earlier versions of this paper.

\end{ack}

%% file: sections/checklist.tex
\section*{Checklist}

%%% BEGIN INSTRUCTIONS %%%
% The checklist follows the references.  Please
% read the checklist guidelines carefully for information on how to answer these
% questions.  For each question, change the default \answerTODO{} to \answerYes{},
% \answerNo{}, or \answerNA{}.  You are strongly encouraged to include a {\bf
% justification to your answer}, either by referencing the appropriate section of
% your paper or providing a brief inline description.  For example:
% \begin{itemize}
%   \item Did you include the license to the code and datasets? \answerYes{See Section~\ref{gen_inst}.}
%   \item Did you include the license to the code and datasets? \answerNo{The code and the data are proprietary.}
%   \item Did you include the license to the code and datasets? \answerNA{}
% \end{itemize}
% Please do not modify the questions and only use the provided macros for your
% answers.  Note that the Checklist section does not count towards the page
% limit.  In your paper, please delete this instructions block and only keep the
% Checklist section heading above along with the questions/answers below.
%%% END INSTRUCTIONS %%%

\begin{enumerate}

\item For all authors...
\begin{enumerate}
  \item Do the main claims made in the abstract and introduction accurately reflect the paper's contributions and scope?
    \answerYes{}
  \item Did you describe the limitations of your work?
    \answerYes{}
  \item Did you discuss any potential negative societal impacts of your work?
    \answerNA{} This work will not be likely to have negative societal impact.
  \item Have you read the ethics review guidelines and ensured that your paper conforms to them?
    \answerYes{}
\end{enumerate}

\item If you are including theoretical results...
\begin{enumerate}
  \item Did you state the full set of assumptions of all theoretical results?
    \answerYes{}
        \item Did you include complete proofs of all theoretical results?
    \answerYes{} Due to the page limits, the complete proofs are included in the supplementary materials.
\end{enumerate}

\item If you ran experiments...
\begin{enumerate}
  \item Did you include the code, data, and instructions needed to reproduce the main experimental results (either in the supplemental material or as a URL)?
    \answerYes{} Code and instructions are included in the supplementary materials. Datasets can be MNIST and FashionMNIST, which can be downloaded online.
  \item Did you specify all the training details (e.g., data splits, hyperparameters, how they were chosen)?
    \answerYes{} The datasplit is default to each dataset. Other details are included in the paper.
        \item Did you report error bars (e.g., with respect to the random seed after running experiments multiple times)?
    \answerNo{} Due to the time limit, we did not include multiple runs for each experiment.
        \item Did you include the total amount of compute and the type of resources used (e.g., type of GPUs, internal cluster, or cloud provider)?
    \answerYes{}
\end{enumerate}

\item If you are using existing assets (e.g., code, data, models) or curating/releasing new assets...
\begin{enumerate}
  \item If your work uses existing assets, did you cite the creators?
    \answerNA{} The datasets are public. We cited the source of the data.
  \item Did you mention the license of the assets?
    \answerNA{}
  \item Did you include any new assets either in the supplemental material or as a URL?
    \answerNA{}
  \item Did you discuss whether and how consent was obtained from people whose data you're using/curating?
    \answerNA{} The datasets are public.
  \item Did you discuss whether the data you are using/curating contains personally identifiable information or offensive content?
    \answerNA{} The datasets involved in this work do not have personal or offensive content. The content is also discussed.
\end{enumerate}

\item If you used crowdsourcing or conducted research with human subjects...
\begin{enumerate}
  \item Did you include the full text of instructions given to participants and screenshots, if applicable?
    \answerNA{}
  \item Did you describe any potential participant risks, with links to Institutional Review Board (IRB) approvals, if applicable?
    \answerNA{}
  \item Did you include the estimated hourly wage paid to participants and the total amount spent on participant compensation?
    \answerNA{}
\end{enumerate}

\end{enumerate}

%% file: appendix_sections/Proofs.tex
\section{Proofs of Lemmas, Statements and Theorems}
% \fixme{Statement 4.2; hyper-parameters; Statement 6.2; Statement 6.3}
% ============ Lemma 4.1 ================
\begin{customlem}{4.1}
% \label{lem:sim_hdc_fail}
No binary HDC can express the following similarity matrix 
\[
\mat{M} = \begin{pmatrix}
        1 & -\frac{1}{2} & -\frac{1}{2} \\
        -\frac{1}{2} & 1 & -\frac{1}{2} \\
        -\frac{1}{2} & -\frac{1}{2} & 1 \\
    \end{pmatrix}
    \; .
\]
\end{customlem}
\begin{proof}
There are $n=3$ basic entities, where we have some HDC vectors $\vec{v}_0, \vec{v}_1, \vec{v}_2\in\R^D$ which can be any dimension.
We start from $D=1$ case, with the inner product as the similarity measurement, we can easily enumerate all possible similarity matrices as follows:
\[
    \left(\begin{smallmatrix} 1 & 1 & 1 \\ 1 & 1 & 1 \\ 1 & 1 & 1 \end{smallmatrix}\right), \;
    \left(\begin{smallmatrix} 1 & -1 & 1 \\ -1 & 1 & -1 \\ 1 & -1 & 1 \end{smallmatrix}\right), \;
    \left(\begin{smallmatrix} 1 & 1 & -1 \\ 1 & 1 & -1 \\ -1 & -1 & 1 \end{smallmatrix}\right), \;
    \left(\begin{smallmatrix} 1 & -1 & -1 \\ -1 & 1 & 1 \\ -1 & 1 & 1 \end{smallmatrix}\right).
\]
When $D>1$, note that 
\[
\textstyle
\simi(\vec{v}_i, \vec{v}_j) = \sum_{k=1}^{D} \simi(\vec{v}_{ik}, \vec{v}_{jk})/D\; ,
\]
which indicates that all possible similarity matrices must reside in the convex hull of the similarity matrices enumerated above because $D$ can be of any dimension. Easy to verify that this convex hell does not contain $\mat{M}$: thus no binary HDC can achieve it. %\tao{characterize the expressivity for HDC.}
% Furthermore, $\vec{v}_1+\vec{v}_2+\vec{v}_3$ cannot be a zero vector; otherwise, consider 
% \[
% 0 = \mathcal{S}(\vec{v}_i,\vec{v}_1+\vec{v}_2+\vec{v}_3) = 1 + \sum_{j\neq i}\mathcal{S}(\vec{v}_i,\vec{v}_{j}),
% \]
% due to symmetry, one can easily derive that $\mathcal{S}(\vec{v}_i,\vec{v}_{j})=-1/2$ for $i\neq j$, then $\mat{M}$ is achieved, a contradiction.
\end{proof}

% ============ Statement 4.1 ================
\begin{customstate}{4.1}
% \label{thm:task_hdc_fail}
% Binary HDC of any dimension cannot learn the following task via bundling method because all class representatives will be the same, unless $\mat{M}$ can be expressed in the HDC system:  \tao{separate to two statements, follows this one}

% let $\mathcal{X}=\{0,1,2\}, x\sim\text{Uniform}(\mathcal{X})$, set
% \[
% y(x) = \left \{ 
% \begin{array}{lr}
%      x & w.p.~~\frac{1}{3}+p  \\
%      x+1, x+2 \bmod 3 & w.p.~~\frac{1}{3}-\frac{p}{2}
% \end{array}
% \right .
% \]
% where $x$ is a scalar input, and $y(x)$ is its ground truth label. $p$ is an arbitrarily small positive number. 

Binary HDC cannot learn the following task.

Consider a supervised learning task with input example set $\mathcal{X}=\{0,1,2\}$, output label set $\mathcal{Y} = \mathcal{X}$, and source distribution
\[
\mathcal{P}(x,y) = \begin{cases} 1/9 + 2p & x = y \\ 1/9 - p & x \ne y \end{cases}
\]
for some small positive number $p$. 
\end{customstate}

\begin{proof}
Let $\phi:\mathcal{X} \rightarrow \{-1,1\}^D$ be any binary HDC encoding, and extend $\phi(x)=\phi(x\bmod 3)$ when $x>3$. Given a class $\hat{y}$, we can then compute the class representative $c_{\hat{y}}$ as 
\[
\begin{split}
    \vec{c}_{\hat{y}} &= \bigoplus_{x:y(x)=\hat{y}} \phi(x) = \sgn(\expect[\phi(x)|\hat{y}])\\ &=\sgn[(\frac{1}{3}-3p)(\phi({\hat{y}+1})+ \phi({\hat{y}+2}))+(\frac{1}{3}+6p)\phi(\hat{y})]\\
    &=\sgn[(\frac{1}{3}-3p)(\phi(\hat{y})+\phi(\hat{y}+1)+\phi(\hat{y}+2))+9p\phi(\hat{y})]\; .
\end{split}
\]
Note that $\phi(\hat{y})+\phi(\hat{y}+1)+\phi(\hat{y}+2)$ cannot be a zero vector; otherwise,  
\[
\begin{split}
    0 &=\mathcal{S}(\phi(\hat{y}),\phi(\hat{y})+\phi(\hat{y}+1)+\phi(\hat{y}+2))\\
    &= 1 + \sum_{j\neq \hat{y}}\mathcal{S}(\phi(\hat{y}),\phi(j)),
\end{split}
\]
due to symmetry, one can easily derive that $\mathcal{S}(\phi(i),\phi(j))=-1/2$ for $i\neq j$, then $\mat{M}$ is achieved, a contradiction to Lemma \ref{lem:sim_hdc_fail}.

Since $p$ is small positive number, then the sign of $\expect[\phi(x)|\hat{y}]$ is dominated by the first term $\phi(\hat{y})+\phi(\hat{y}+1)+\phi(\hat{y}+2)$. Hence, the class representative of class $\hat{y}$ is computed as 
\[
\begin{split}
    \vec{c}_{\hat{y}} &= \bigoplus_{x:y(x)=\hat{y}} \phi(x) = \sgn(\expect[\phi(x)|\hat{y}]) \\
    &= \sgn(\phi(\hat{y})+\phi(\hat{y}+1)+\phi(\hat{y}+2)),
\end{split}
\]
which is same for each class, i.e., binary HDC fails to learn this simple task. 

On the other hand, for any VSA that can express $\mat{M}$ with $\vec{v}_0, \vec{v}_1, \vec{v}_2$, set $\phi(x)=\vec{v}_{x}$. We can compute the class representative $c_{\hat{y}}$ as
\[
\begin{split}
    \vec{c}_{\hat{y}} &= \bigoplus_{x:y(x)=\hat{y}} \phi(x) =\arg \max_{z} \langle z, \expect[\phi(x)|\hat{y}] \rangle \\
    &= \arg \max_{z} \langle z, (\frac{1}{3}-3p)(\phi(\hat{y})+\phi(\hat{y}+1)+\phi(\hat{y}+2))+9p\phi(\hat{y}) \rangle \\
    &= \arg \max_{z} \langle z, 9p\phi(\hat{y}) \rangle = \phi(\hat{y}).
\end{split}
\]
The class representative of class $\hat{y}$ will be
\[
c_{\hat{y}} = \sgn(\expect[\phi(x)|\hat{y}]) = \sgn(\frac{3p}{2}\phi(\hat{y})) = \phi(\hat{y}),
\]
which gives a Bayes optimal classifier, outputs the most probable class and proves Statement 4.2.
\end{proof}

% ============ Statement 4.2 ================
\begin{customstate}{4.2}
% \label{thm:task_hdc_succeed}
Any VSA (formalized in Definition $2$) that can express $\mat{M}_{\textup{Lemma \ref{lem:sim_hdc_fail}}}$ can learn this task.
\end{customstate}

% ============ Lemma 4.2 ================
\begin{customlem}{4.2}
% \label{lemma:classichdc}
Let $u_1, u_2, \ldots, u_K$ be binary vectors sampled coordinate-wise independently at random, where each coordinate of $u_i$ has the same probability $p_i$ of being $1$.
Let $v_0$, $v_1$, and $v_2$ be vectors that result from some composition of binding and permutation acting on $u_1, \ldots, u_K$, and let $\mat{M} \in \R^{3 \times 3}$ be their similarity matrix, such that $\mat{M}_{ij} = \simi(v_i, v_j)$. Then
\[
\left\| \expect[\mat{M}] - \begin{pmatrix}
        1 & -\frac{1}{3} & -\frac{1}{3} \\
        -\frac{1}{3} & 1 & -\frac{1}{3} \\
        -\frac{1}{3} & -\frac{1}{3} & 1 \\
    \end{pmatrix} \right\|_F \ge \frac{\sqrt{2}}{3},
\]
but this target matrix \emph{is} expressible by some binary HDC.
\end{customlem}
\begin{proof}
It is straightforward to show that %\textcolor{red}{[add proof]}
for some $x, y, z \in [-1,1]$
\[
\mathbf{E}[\mat{M}] = \begin{pmatrix}
        1 & xy & xz \\
        xy & 1 & yz \\
        xz & yz & 1 \\
    \end{pmatrix}.
\]
But since $(xy) \cdot (xz) \cdot (yz) = x^2 y^2 z^2$ is a square number, it follows that the upper-triangular elements cannot all be negative. At least one of them must be non-negative, from which the result immediately follows. A binary HDC that achieves this matrix is: $(-1,1,1)$, $(1,-1,1)$, $(1,1,-1)$.
\end{proof}

% ============ Lemma 5.1 ================
\begin{customlem}{5.1}
Suppose that $X$ and $Y$ are jointly Gaussian zero-mean unit-variance random variables. Then
\[
\expect[\sgn(X)\sgn(Y)] = \frac{2}{\pi} \arcsin\left( \expect[X Y ] \right)
\]
% and with covariance $\expect{X Y} = $
% Assume $\vec{x}=(x_1,\ldots, x_n)^\top$ with each element drawn from $\mathcal{N}(0,1)$, let $u=\vec{a}^\top\vec{x}, v=\vec{b}^\top\vec{x}$, then 
% \[
% \expect[\sgn(u)\sgn(v)] = 1-\frac{2}{\pi}\arccos(\theta),
% \]
% where $\theta=\arccos(\frac{\vec{a}^\top\vec{b}}{\|\vec{a}\|\cdot\|\vec{b}\|})$.
\end{customlem}
% \begin{proof}
% Note that 
% \[
% \begin{split}
%     &\expect[\sgn(u)\sgn(v)] = 1-\frac{\pi}{2}\arccos(\rho)\\
%     =&\prob(u>0,v>0)-\prob(u>0,v<0)\\
%     &+\prob(u<0,v<0)-\prob(u<0,v>0) \\
%     =&\prob(u>0,v>0)-(\prob(u>0) - \prob(u>0,v>0))\\
%     &+\prob(u<0,v<0)-(\prob(u<0)-\prob(u<0,v>0)) \\
%     =&2\prob(u>0,v>0)+2\prob(u<0,v<0)-\prob(u>0)\\
%     &-\prob(u<0)\\
%     =&2(\prob(u>0,v>0)+\prob(u<0,v<0))-1\\
%     =&2(\frac{\pi-\theta}{2\pi}+\frac{\pi-\theta}{2\pi})-1\\
%     =&1-\frac{2\theta}{\pi}\;.
% \end{split}
% \]
% \end{proof}
\begin{proof}
Without loss of generality let $U \sim \mathcal{N}(0,I)$ be a standard Gaussian over $\R^2$, and suppose that $X = a^T U$, $Y = b^T U$ for some vectors $a,b \in \R^2$ with $\|a\|=\|b\|=1$ and $a^T b = \expect[XY]$. Here, a geometric argument shows that $\prob(X \ge 0 \land Y \le 0) = \prob(a^T U \ge 0 \land b^T U \le 0) = \theta/(2 \pi)$, where $\theta$ is the angle between $a$ and $b$. An analogous analysis of the other three cases, combined with some straightforward trigonometry, proves the lemma.
\end{proof}

\newcommand{\trace}[1]{\operatorname{tr}\left( #1 \right)}

% ============ Theorem 6.1 ================
\begin{customthm}{6.1}
Let $(G, \otimes)$ be a finite group, and let $X$ denote the set of its non-trivial irreducible characters.
Let $\alpha: X \rightarrow \R_{\ge 0}$ be some function that assigns a non-negative weight to each of the characters. Then, if we set $\mathcal{S}$ to be
\[
    \mathcal{S}(g, h) = \frac{\sum_{\chi \in X} \alpha(\chi) \cdot \operatorname{Re}(\chi(g^{-1} \otimes h))}{\sum_{\chi \in X} \alpha(\chi) \cdot \chi(\mathbf{1})},
\]
where the inverse and unit $\mathbf{1}$ are those of the group $(G, \otimes)$, and define bundling $\oplus$ as given in the definition of a group VSA, then $(G, \otimes, \mathcal{S}, \oplus)$ is a finite group VSA. Any finite group VSA can be constructed in this way.

If in this construction $\alpha$ is supported on only one character $\chi$, i.e. $\mathcal{S}(g, h) =\operatorname{Re}(\chi(g^{-1} \otimes h))/ \chi(\mathbf{1})$, then the VSA will have the product property.
\end{customthm}
\begin{proof}
The first part of this theorem is a direct consequence of the following more technically-stated theorem. 

The second part follows directly from the fact that if $\phi$ is an irreducible representation of a finite group $G$, and $A$ is the set of automorphisms of $G$, then for any $g \in G$,
\[
    \frac{1}{|A|} \sum_{a \in A} \phi(a(g)) = cI
\]
for some scalar $c$. In particular, this means that if $\chi$ is the corresponding character, then for any $g,h \in G$,
\begin{align*}
    \frac{1}{|A|} \sum_{a \in A} \chi(h^{-1} a(g))
    &=
    \frac{1}{|A|} \sum_{a \in A} \trace{\phi(h^{-1}) \phi(a(g))}
    \\&=
    \trace{\phi(h^{-1}) \frac{1}{|A|} \sum_{a \in A} \phi(a(g))}
    \\&=
    \trace{\phi(h^{-1}) \cdot cII}
    \\&=
    c \cdot \chi(h^{-1}).
\end{align*}
Substituting $h = 1$ yields that $c = \operatorname{Re}(\chi(g)) / \chi(1)$ (since $\chi$ must also be preserved by automorphisms up to complex conjugation),
which immediately implies what we wanted to prove.
\end{proof}

\begin{theorem}[Representation theorem for group VSAs]
Suppose that we have a finite group $G$ equipped with a similarity measure denoted $\langle \cdot | \cdot \rangle_G$.
Let $X$ denote the set of non-trivial irreducible characters of $G$ (i.e. $X$ is the character table of $G$ excluding the top row).
Then there exists a unique function $\alpha: X \rightarrow \R_{\ge 0}$ such that $\alpha(\bar \chi) = \alpha(\chi)$ for all $\chi \in X$, $\sum_{\chi \in X} \alpha(\chi) \cdot \chi(1) = 1$, and
\begin{equation}
    \label{eqnDefSim}
    \langle \cdot | \cdot \rangle_G
    =
    \sum_{\chi \in X} \alpha(\chi) \cdot \chi(g^{-1} h).
\end{equation}
Conversely, for any function $\alpha$ of this type, if we define $\langle \cdot | \cdot \rangle$ according to (\ref{eqnDefSim}), then $\langle \cdot | \cdot \rangle$ will be a similarity measure for $G$.

Additionally, if we define $M$ as the matrix such that $M_{gh} = \langle g | h \rangle_G$, and $d$ is the rank of $M$, then there exists some positive integer $K$ and positive integers $d_1, d_2, \ldots, d_K$ such that $\sum_{k=1}^K d_K = 1$, and there exists a $|G|$-dimensional subspace $\mathcal{A}$ of $\R^{d_1 \times d_1} \times \R^{d_2 \times d_2} \times \cdots \times \R^{d_K \times d_K}$ and function $\phi: G \rightarrow \mathcal{A}$ such that for all $g, h \in G$,
\begin{align*}
    \phi(g) \phi(h) &= \phi(gh) \\
    \phi(g)^{-1} = \phi(g)^T &= \phi(g^{-1}) \\
    \phi(1) &= \phi(I),
\end{align*}
where multiplication and transposition in $\mathcal{A}$ is done component-wise on the $K$ components (each of which is a matrix), and such multiplication preserves $\mathcal{A}$.
Also, there exist some positive scalars $\beta_1, \beta_2, \ldots, \beta_K$ such that if we define an inner product on $\mathcal{A}$ as
\[
    \langle (x_1, x_2, \ldots, x_K), (y_1, y_2, \ldots, y_K) \rangle_{\mathcal{A}} = \sum_{k=1}^K \beta_k \trace{x_k^T y_k},
\]
where here each $x_k \in \R^{d_k \times d_k}$, then
\[
    \langle g | h \rangle_G = \langle \phi(g) | \phi(h) \rangle_{\mathcal{A}}.
\]
\end{theorem}
\begin{proof}
Let $M$ be the matrix described in the theorem statement, such that $M_{gh} = \langle g | h \rangle_G$. Observe that $M$ must be symmetric and positive semidefinite (by the properties of the similarity measure).
For some $f \in G$, let $B_f$ denote the matrix
\[
    B_f = \sum_{g \in G} e_{fg} e_g^T,
\]
where $e_g$ is the unit basis element associated with $g$.
Observe that $B_f^T = B_f^{-1} = B_{f^{-1}}$, and that $B_f$ commutes with $M$, because
\begin{align*}
    e_h^T M B_f e_g
    &=
    e_h^T M e_{fg} \\
    &=
    \rho(h^{-1} f g) \\
    &=
    \rho((f^{-1} h)^{-1} g) \\
    &=
    e_{f^{-1} h}^T M e_g \\
    &=
    \left( B_{f^{-1}} e_{h} \right)^T M e_g \\
    &=
    e_{h}^T B_{f^{-1}}^T M e_g \\
    &=
    e_{h}^T B_f M e_g.
\end{align*}
That is, $M B_f = B_f M$.
Since $B_f$ commutes with $M$, it also must commute with any polynomial of $M$. In particular, if the nonzero eigendecomposition of $M$ is
\[
    M = \sum_{k=1}^K \lambda_k V_k,
\]
where each $\lambda_k > 0$ is distinct, and $V_k^2 = V_k$ is a symmetric projection matrix onto the associated eigenspace, then since $V_k$ can be expressed as a polynomial in $M$, it must also commute with $B_f$ for all $f$.
The same thing will be true for $C_f$ defined as
\[
    C_f = \sum_{g \in G} e_{gf} e_g^T.
\]
These results together show that for any $f, g, h \in G$,
\[
    e_h^T V_k e_g = e_{fh}^T V_k e_{fg} = e_{hf}^T V_k e_{gf}.
\]

Now, let $d_k$ denote the rank of $V_k$ (the multiplicity of the eigenvalue $\lambda_k$ in $M$).
So, there must exist some matrix $W_k \in \R^{|G| \times d_k}$ such that $V_k = W_k W_k^T$ and $W_k^T W_k = I$.
For any $g \in G$, let $U_k(g)$ denote the matrix in $\R^{d_k \times d_k}$
\[
    U_k(g)= W_k^T B_g W_k.
\]
Observe that $U_k(1) = I$, all the $U_k(g)$ matrices are orthogonal, and
\begin{align*}
    U_k(g) U_k(h)
    &=
    W_k^T B_g W_k W_k^T B_h W_k \\
    &=
    W_k^T B_g V_k B_h W_k \\
    &=
    W_k^T B_g B_h V_k W_k \\
    &=
    W_k^T B_{gh} W_k \\
    &=
    U_k(gh).
\end{align*}
That is, $U_k$ is a representation of the group $G$.
Observe that the trace of $U_k(g)$ is
\begin{align*}
    \trace{U_k(g)}
    &=
    \trace{W_k^T B_g W_k}
    \\&=
    \trace{W_k W_k^T B_g}
    \\&=
    \trace{V_k B_g}
    \\&=
    \sum_{h \in G} e_h^T V_k B_g e_h
    \\&=
    \sum_{h \in G} e_h^T V_k e_{gh}
    \\&=
    \sum_{h \in G} e_1^T V_k e_g
    \\&=
    |G| \cdot e_1^T V_k e_g.
\end{align*}
In particular, this means that $g \mapsto |G| \cdot e_1^T V_k e_g$ is a character of $G$.
As a character, it must be a sum of irreducible characters of $G$, and since it is real, it must place the same weight on complex-conjugate characters.
It follows that
$g \mapsto \sum_{k=1}^K \lambda_k e_1^T V_k e_g$ must be a non-negative scaled sum of irreducible characters of $G$ that places the same weight on complex-conjugate characters. But, this function is just $g \mapsto e_1^T M e_g$, which is just $\rho$.
So, $\rho$ must be a non-negative sum of the irreducible characters of $G$.
The fact that this scaling is unique follows from the fact that the characters are linearly independent; the fact that this scaling only contains non-trivial characters follows from the average-dissimilarity property.

We then construct an algebra $\mathcal{A}$ as follows.
Let $\phi(g): G \rightarrow \R^{d_1 \times d_1} \times \cdots \times R_{d_K \times d_K}$ be defined as
\[
    \phi(g) = (U_1(g), U_2(g), \ldots, U_K(g)), 
\]
and let the inner product scalars be
\[
    \beta_k = \frac{\lambda_k}{|G|}.
\]
Then
\begin{align*}
    \sum_{k=1}^K \beta_k \trace{U_k(h)^T U_k(g)}
    &=
    \sum_{k=1}^K \beta_k \trace{U_k(h^{-1}) U_k(g)}
    \\&=
    \sum_{k=1}^K \beta_k \trace{U_k(h^{-1} g)}
    \\&=
    e_1^T M e_{h^{-1} g}
    \\&=
    \rho(h^{-1} g)
    \\&=
    \langle h | g \rangle_G
\end{align*}
as desired.
To finish the construction, let $\mathcal{A}$ be the algebra spanned by $\{ \phi(g) \mid g \in G \}$.
Observe that this must be closed under multiplication and transposition because $G$ is closed under multiplication and inversion.
\end{proof}

% ============ Statement 6.1 ================
\begin{customstate}{6.1}
Let $\mat{M}$ be similarity matrices expressible by a finite group VSA. Then there exists a finite group VSA that has the product property and can also achieve $\mat{M}$.
\end{customstate}
\begin{proof}
Suppose the first VSA's group is $G$ and has irreducible characters $\chi_1, \chi_2, \ldots, \chi_k$. Then the group $G^k$ consisting of the direct product of $k$ copies of the group $G$, together with a similarity function $\simi((x_1, \ldots, x_k), (y_1, \ldots, y_k)) \propto \prod_{i=1}^k \chi_i(x_i^{-1} \otimes y_i)$ will both have the product property (as its similarity matrix is proportional to a single character) and can express any similarity matrix the first VSA can.
\end{proof}

% ============ Statement 6.2 ??? ================
\begin{customstate}{6.2}
% \label{stmt:abelian}
Any similarity matrix $\mat{M}$ that can be expressed by a finite Abelian group VSA can be expressed by the unit-cycle VSA ($G = \{z \in C \mid |z| = 1\}$, $x \otimes y = xy$, $\simi(x,y) = \operatorname{Re}(x^* y)$).
\end{customstate}
\begin{proof}
Note the fact that all irreducible representations of finite Abelian groups are 1-dimensional. Then the results follows immediately with any irreducible representation as the mapping from a finite Abelian group to $G = \{z \in C \mid |z| = 1\}$.
\end{proof}

\begin{customstate}{6.3}
% \label{stmt:nonabelian}
There exists a similarity matrix $\mat{M}$ that can be expressed by a VSA over the (non-Abelian) binary icosahedral group, but not by the unit-cycle VSA.
\end{customstate}
% \begin{customthm}{6.2}
% There exists similarity matrices that can be expressed by the binary icosahedral group, but not by $\C$.
% \end{customthm}
\begin{proof}
%\tao{move this proof to appendix}
Consider the binary icosahedral group expressed as a subset of the quaternions. This group consists of 120 elements which are placed at the vertices of a 600-cell inscribed in the unit 3-sphere. Consider an arbitrary sequence of 60 of these elements containing exactly one of $\{x, -x\}$ for each $x$ in the binary icosahedral group: i.e. we select exactly one of each pair of antipodal points in the group.  This sequence of 60 points has a similarity matrix $M \in \mathbb{R}^{60 \times 60}$. It is easy to check that the absolute values of the entries of this matrix lie in $\{0, \frac{1}{2 \phi}, \frac{1}{2}, \frac{\phi}{2}, 1\}$, where $\phi = \frac{1 + \sqrt{5}}{2}$ is the golden ratio. Define the matrix $A \in \mathbb{R}^{60 \times 60}$ such that $A_{ij} = \pm 1$ if $M_{ij} = \pm \frac{\phi}{2}$ and $A_{ij} = 0$ otherwise. Consider the optimization problem to maximize $\operatorname{tr}(AZ)$ over positive semidefinite matrices $Z \in \mathbb{R}^{60 \times 60}$ subject to the constraint that the diagonal of $Z$ is all-ones, i.e. $Z_{ii} = 1$. It is easy to check numerically that the only solution to this optimization problem is $Z = M$. Also observe that $M$ has rank greater than $2$.

Now, suppose that there existed some representation of $M$ using vectors with entries in the unit circle in $\mathbb{C}$. For this to hold, it would need to be the case that $M$ is in the convex combination of the similarity matrices generated by those entries, each of which must be of rank $2$. But, this is impossible, since (1) none of those matrices can be equal to $M$, and as such (2) any such matrix $M_C$ will have $\operatorname{tr}(M_C A) < \operatorname{tr}(M A)$. This shows that this particular matrix can't be represented by hypervectors with unit-absolute-value entries in $\mathbb{C}$.
\end{proof}

% ============ Statement 6.3 ??? ================

\section{Calculation of $\theta$ in Section 7}
% \tao{missing proofs @Yichi}
We can theoretically calculate the angle $\theta$ between the class vector $s_c$ and a randomly selected hypervector $t_{i}$ in the set $\mathbb{T}_{c}$.
$$ \text{cos} \theta = \frac{s_{c} \cdot v_{j}}{\left \| s_{c} \right \|  \left \| v_{j} \right \|} = \frac{2 \cdot q - D}{D} $$
% $$ \text{cos} \theta = \frac{s_{c} \cdot v_{j}}{\left \| s_{c} \right \|  \left \| v_{j} \right \|} = \frac{2 \cdot q - D}{D} = \frac{2 \cdot D \cdot p - D}{D} = 2 \cdot p - 1 $$
$D$ is the dimension of $t_{i}$. $q$ is the number of elements that have the same sign in $s_c$ and $t_i$.
Since 
\[s_{c} = \text{sgn} \left( \bigoplus_{j \in \mathbb{T}_{c}} t_j \right) = \text{sgn} \left( t_{i} + \sum_{j=1, j \neq i}^{2k+1} t_{j} \right)\;, 
\]
$q$ is proportional to the probability $p_k$ of the sign of an entry in $t_i$ will be flipped after adding the term $\sum_{j=1, j \neq i}^{2k+1} t_{j}$ (the other $2k$ vectors) to it.
Obviously $q = D \cdot p_{k}$ as each entry in $t_i$ is independent.

In order to avoid flipping the sign of an entry, there should be at least $k$ entries out of $2k$ that have the same sign, so the probability $p_k$ can be calculated by
$$ p_{k} = \frac{\binom{2k}{k}+\binom{2k}{k+1}+\cdots+\binom{2k}{2k}}{2^{2k}} = \frac{1+\frac{1}{2^{2k}}\binom{2k}{k}}{2} $$
Plug the expression of $p_k$ into $\text{cos} \theta$, we can get
$$ \theta_{2k+1} = \text{cos}^{-1}(\frac{1}{2^{2k}} \cdot \binom{2k}{k}) $$

Importantly, $p_k$ is monotonic.
\begin{proof}
\begin{align*} 
p_{k+1} &= \frac{1 + \frac{1}{2^{2k}} \cdot \frac{1}{4} \cdot {2k+2 \choose k+1} }{2} \\
        &= \frac{1 + \frac{1}{2^{2k}} \cdot \frac{1}{4} \cdot [{2k+1 \choose k+1} + {2k+1 \choose k}] }{2} \\
        &= \frac{1 + \frac{1}{2^{2k}} \cdot \frac{1}{4} \cdot [{2k \choose k+1} + {2k \choose k} + {2k \choose k} + {2k \choose k-1}] }{2} \\
        &< \frac{1 + \frac{1}{2^{2k}} \cdot \frac{1}{4} \cdot 4 \cdot {2k \choose k} }{2} \\
        &= p_{k}
\end{align*}
\end{proof}
This means that the more vectors we bundle together, the closer $\theta$ is to 90 degrees.

%% file: appendix_sections/GroupLearning.tex
\section{Learning with Group VSA}
For a cyclic group VSA model, similar as the binary HDC case, we initialize a linear model with weights $\mat{W}$ of size $\# \operatorname{class} \times D$, where each element belongs to $G=\mathbb{Z}/n\mathbb{Z}=\{0, 1, \cdots, n-1\}$. Inputs to this classifier are encoded hypervectors $v\in G^D$, the model computes per-class similarities with defined similarity function:
\[
\simi(x,y) = \langle \psi(x), \psi(y) \rangle = \cos(2\pi(x-y)/n), \forall x,y\in G,
\]
which extends to higher dimensional space via 
\[
\mathcal{S}([x_1, \ldots, x_D],[y_1, \ldots, y_D]) = \frac{1}{D} \sum_{i=1}^D \mathcal{S}(x_i, y_i).
\]
We calculate the cross-entropy loss between the classifier outputs and the labels, and do back propagation using high precision numbers such as $16$-bit floats. At each optimizer step, SGD optimizer moves $\mat{W}$ away from $G^{\# \operatorname{class} \times D}$, we pull it back with (Fast)-Round operation to finish current step before entering next step. Similar as the binary case, the inference cost remains the same as the bundling method.